\documentclass{article}

\PassOptionsToPackage{numbers, sort&compress}{natbib}

 \usepackage[main, final]{neurips_2026}

\usepackage[utf8]{inputenc} 
\usepackage[T1]{fontenc}    
\usepackage{url}            
\usepackage{booktabs}       
\usepackage{amsfonts}       
\usepackage{nicefrac}       
\usepackage{microtype}      
\usepackage{xcolor}         

\usepackage{xspace}
\usepackage{amssymb}
\usepackage{caption}

\usepackage{amsmath}

\usepackage{algorithm}
\usepackage{algpseudocode}

\usepackage{multirow}
\usepackage{wrapfig}
\usepackage{makecell}
\usepackage[table]{xcolor}

\usepackage[normalem]{ulem}

\usepackage{graphicx}

\usepackage{uoftcolors}

\usepackage[pagebackref]{hyperref}

\hypersetup{colorlinks=true,allcolors=uoftsecondaryblue}

\title{Squeezing Capacity from Multimodal Large Language Models for Subject-driven Generation}

\author{
\vspace{2mm} 
        Shuhong Zheng$^{1,2}$\hspace{6mm}
        Aashish Kumar Misraa$^{3*}$ \hspace{6mm}
        Yu-Teng Li$^{3*}$
        \\
        \textbf{
        Yu-Jhe Li$^{4{\P}*\dagger}$ \hspace{6mm}
         Igor Gilitschenski$^{1,2\dagger}$
    }
    \vspace{2mm}
    \\
    $^{1}$University of Toronto
    \hspace{3mm}  
    $^{2}$Vector Institute
    \hspace{3mm}  
    $^{3}$Adobe Research
    \hspace{3mm}
    $^{4}$Google
    }

\begin{document}

\maketitle

\renewcommand{\thefootnote}{$\P$}
\footnotetext{Work done in Adobe}
\renewcommand{\thefootnote}{\arabic{footnote}}

\renewcommand{\thefootnote}{*}
\footnotetext{Equal Contribution}
\renewcommand{\thefootnote}{\arabic{footnote}}

\renewcommand{\thefootnote}{$\dagger$}
\footnotetext{Joint Advising}
\renewcommand{\thefootnote}{\arabic{footnote}}



\begin{abstract}

Subject-driven image generation aims to synthesize new images that preserve the identity of the given subject while following textual instructions. Existing approaches often encode text and reference images separately. This limits cross-modal reasoning abilities and causes copy-paste artifacts. Recent frameworks that connect multimodal models and diffusion models improve instruction following, but largely overlook identity preservation. To address these limitations, we condition diffusion models on Multimodal Large Language Models (MLLMs) that jointly encode text and reference images, and augment it with VAE‑based identity conditioning. A novel Dual Layer Aggregation (DLA) module is designed to aggregate multi‑level MLLM features for optimal conditioning, and a multi‑stage denoising strategy is applied to progressively balance the semantic information from MLLM and fine‑detail identity from VAE during inference. Extensive experiments demonstrate that our approach harmonizes multimodal understanding with identity preservation, mitigates copy-paste issues, and achieves superior performance regarding human preference on subject-driven image generation. Our project website is available at \url{https://zsh2000.github.io/squeeze-mllm-subject-gen/}.

\end{abstract}
\section{Introduction}
\label{sec:intro}




Subject-driven image generation aims to synthesize new content while preserving the visual identity of a specific subject. Early approaches~\cite{voynov2023pextendedtextualconditioning, liu2023conesconceptneuronsdiffusion, dong2025dreamartistcontrollableoneshottexttoimage, kumari2022customdiffusion, wei2023elite, breakascene_sa23, chen2024disenbooth, liu2023customizable_cones2}, such as DreamBooth~\cite{ruiz2023dreambooth} and Textual Inversion~\cite{gal2022image}, personalize pretrained diffusion models via per-subject fine-tuning, achieving strong identity fidelity at the cost of scalability. Subsequent works~\cite{dong2024how, shi2023instantboothpersonalizedtexttoimagegeneration, huang2024realcustom, realcustom++, subjectdiffusion_sig24, zhang2024ssrencoder, purushwalkam2024bootpigbootstrappingzeroshotpersonalized} adopt reference-image conditioning to avoid retraining, where models like IP-Adapter~\cite{ye2023ip} extract subject features at inference time. More recent efforts~\cite{wu2025less, cai2024dsd, ouyang2025consistencycriticcorrectinginconsistencies, fu2025imontageunifiedversatilehighly, ye2025visualawarecotachievinghighfidelity, dong2026echogen} further enhance zero-shot subject generalization through VAE-based (Variational Autoencoder-based~\cite{vae_iclr14}) token conditioning. However, these pipelines still process text and reference images separately, limiting multimodal understanding and often causing copy-paste artifacts (reproducing the reference subject with the identical pose without adaptation to the generated scenarios) or identity drift on complex prompts.

In parallel, multimodal large language models (MLLMs)~\cite{liu2024_llava15, liu2023visual} have demonstrated strong abilities in joint text-image reasoning and structured control~\cite{singhania2025pixelsvlmbasedevaluationidentity}. Systems~\cite{dalva2025canvastoimagecompositionalimagegeneration, singhania2025tamingidentityconsistencyprompt} such as DreamEngine~\cite{chen2025multimodal}, Qwen-Image~\cite{wu2025qwen}, and EasyRef~\cite{zong2024easyref} integrate MLLMs into diffusion decoders to parse interleaved multimodal instructions, enabling more flexible prompt interpretation. Yet, these designs typically rely only on the MLLM's final-layer features (\textit{e.g.,} Qwen-Image, EasyRef), or combine ViT features which contain fine details, with final-layer outputs via scalar mixing (\textit{e.g.,} DreamEngine). These models often neglect fine-grained visual cues which are crucial for identity, thereby leading to suboptimal identity preservation.

In this work, we unify these two directions by introducing an MLLM-driven subject conditioning framework that jointly encodes text and reference images within a shared multimodal space, and enhances ID preservation with VAE conditioning. This joint encoding enables the model to perform multimodal reasoning and coherently preserve subject identity, beyond the representational limits of pure VAE-based encoders. However, this unification is non-trivial due to the different feature structures of text and image tokens in MLLMs. 
The discrepancy between text and image features makes it fundamentally inadequate to directly fuse modalities or rely on a single-layer representation for conditioning. 
To effectively align MLLM embeddings with diffusion features, we design an innovative Dual Layer Aggregation (DLA) mechanism, that adopts layerwise attention pooling to separately aggregate text and visual embeddings. Instead of conditioning solely on the MLLM's final layer feature, the DLA takes the aggregated features from all transformer layers in the MLLM as input, to fully leverage its multimodal prompt understanding capability. We also justify the mechanism of aggregation by analyzing the roles and effectiveness of different layer groups (\textit{i.e.,} early, middle, and late layers) within MLLM in the experimental study.

\begin{wrapfigure}[24]{r}{0.55\textwidth}
    \centering
    \vspace{-2mm}
    \includegraphics[width =\linewidth]{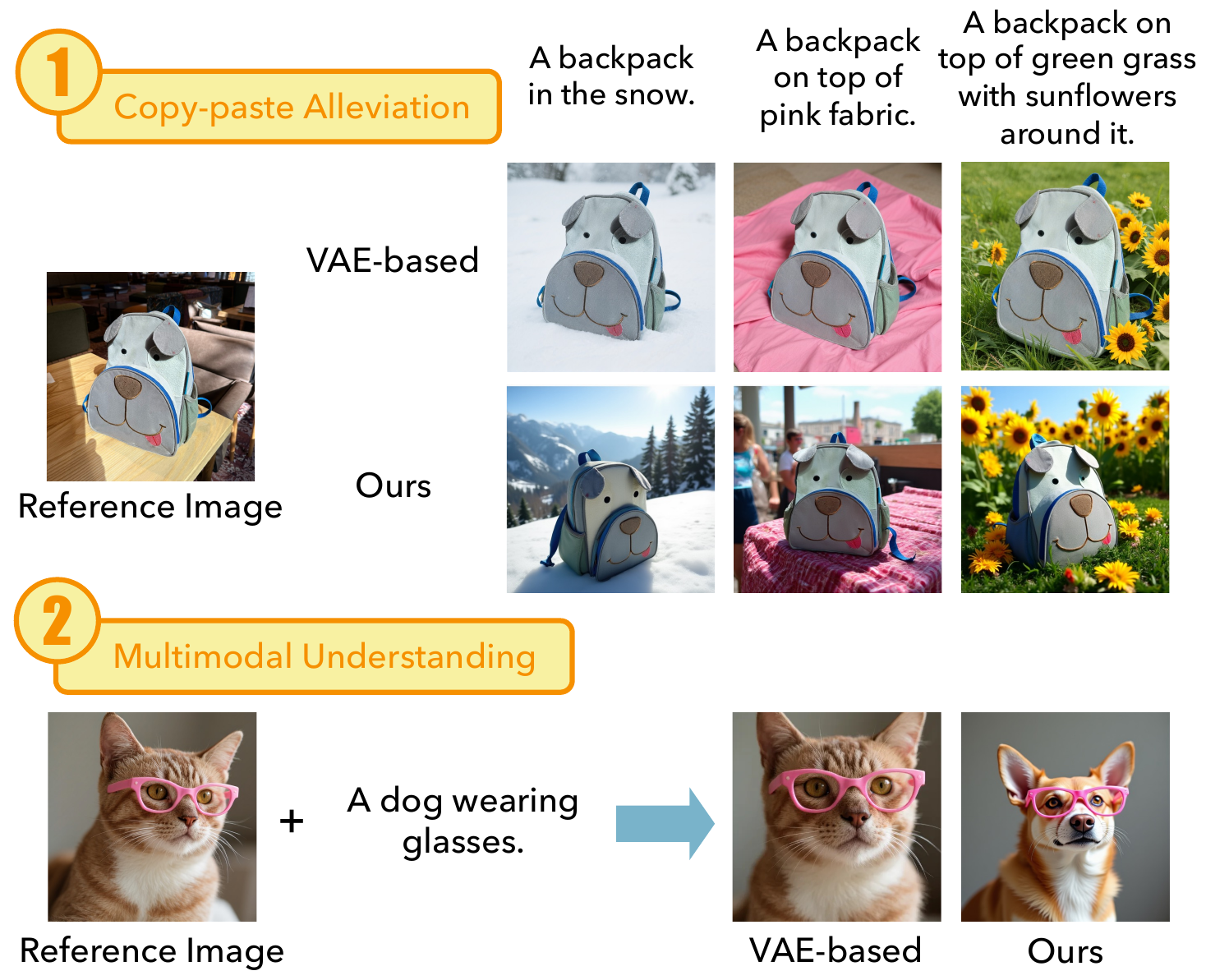}
    \vspace{-5mm}
    \caption{Benefits of leveraging MLLMs for subject-driven generation. MLLMs mitigate the \textit{copy-paste issue} within VAE-based methods, and enables the \textit{multimodal understanding} of the subject-driven generation pipeline by jointly modeling input image and text, while VAE-based methods encode them separately.}
\vspace{-20pt}
    \label{fig:teaser}
\end{wrapfigure}
In addition, directly combining MLLM embeddings with VAE-based identity enhancement can cause embedding conflicts, as both contain overlapping visual representations. To reconcile these signals, a two-stage training strategy is invented to first enable multimodal conditioning from MLLM, before combining the optimization with the high-frequency identity details from VAE features. To further balance the multimodal conditioning from the MLLM and the identity details provided by the VAE, we propose a multi-stage denoising strategy: the diffusion model first denoises under MLLM guidance to establish global semantics, then jointly refines with both modalities, and finally focuses on VAE-conditioned fine details. As shown in Figure~\ref{fig:teaser}, this staged process effectively harmonizes the two embedding sources, alleviating copy-paste artifacts common in VAE-based pipelines, while providing richer reasoning ability and instruction-aware, identity-preserving generation compared to existing frameworks. Our contributions can be summarized as follows: we propose a Dual Layer Aggregation (DLA) module to aggregate text and visual embeddings across MLLM layers for improved conditioning, along with a multi-stage denoising strategy that balances semantic reasoning and fine-grained identity during generation. Also, we provide a detailed analysis of MLLM layer representations and their roles in diffusion conditioning under different fusion strategies. Extensive experiments demonstrate competitive performance in multimodal understanding and identity preservation over prior subject-driven methods.
\section{Related work}
\label{sec:related}




\textbf{Subject-driven generation} focuses on preserving the identity or visual characteristics of a specific subject within the synthesized images. Early optimization-based approaches~\cite{chen2023subjectdriven, han2023svdiff, encoder_based_tog23, domain_agnostic_finetuning_sa23} such as DreamBooth~\cite{ruiz2023dreambooth}, Textual Inversion~\cite{gal2022image}, and LoRA~\cite{hu2022lora} adapt pretrained diffusion models to new identities by introducing subject-specific parameters, but require costly per-subject fine-tuning. To eliminate this need, recent methods employ explicit reference encoders or adapters that extract identity features directly from input images and condition the diffusion process at inference time (\textit{e.g.}, IP-Adapter~\cite{ye2023ip}, BLIP-Diffusion~\cite{li2023blip}). Transformer-based diffusion decoders (DiT) have further incorporated such reference conditioning~\cite{li2025iccustomdiverseimagecustomization, hu2026positionicunifiedpositionidentity} through lightweight modules like IC-LoRA~\cite{huang2024context}. Subsequent research~\cite{huang2025competitionsynergyunlockingreinforcement, li2025editidtrainingfreeeditableid, xing2026lumosx, li2026bindweave, jin2026consisgcpo} enhances facial fidelity~\cite{qian2025layercomposermultihumanpersonalizedgeneration, xu2026withanyone, wu2025multicrafterhighfidelitymultisubjectgeneration, li2025editidv2editableidcustomization, tao2025instantcharacter, wang2025charcomcomposableidentitycontrol}, multi-reference composition~\cite{wang2024ms, jin2025focusdpodynamicpreferenceoptimization, zhang2025creatidesignunifiedmulticonditionaldiffusion, zheng2025freeloraenablingtrainingfreelora, she2025mosaicmultisubjectpersonalizedgeneration, xu2026contextgen, shi2025consistcomposeunifiedmultimodallayout, wang2025psrscalingmultisubjectpersonalized, tarres2026placididentitypreservingmultiobjectcompositing, xu2026hierarchicalconcepttoappearanceguidancemultisubject, wei2026unirefimageeditscalableconsistentmultireference,yang2025hicogenhierarchicalcompositionaltexttoimage, saha2026sigmagen, guo2025musar}, computational efficiency~\cite{jia2023tamingencoderzerofinetuning, li2025dvidisentanglingsemanticvisual, li2026injectmatterstrainingfreespatiallyadaptive, dalva2025lorashop, wang2025dynaipdynamicimageprompt, yang2025echodistillbidirectionalconceptdistillation, yao2025freegraftor, li2024tuning}, and multimodal controllability~\cite{guo2024pulid,xiao2025omnigen,le2025one, jang2024identity, fang2025tbstareditimageeditingpattern, xia2025dreamomniunifiedimagegeneration, xia2025dreamomni2multimodalinstructionbasedediting, wang2025sconebridgingcompositiondistinction, he2026realign, VisualCloze_2025_ICCV,song20253sgenunifiedsubjectstyle}. Recently, UNO~\cite{wu2025less}, UMO~\cite{cheng2025umo}, USO~\cite{wu2025uso}, and DreamO~\cite{mou2025dreamo} achieve zero-shot generation conditioned by multiple images leveraging VAE-based token conditioning. However, these identity-preserving and control-oriented pipelines remain largely decoupled from large multimodal language models (MLLMs), lacking the semantic reasoning and contextual understanding necessary for flexible, instruction-aware identity control. 


\noindent\textbf{Text-to-image (T2I) generation} has rapidly advanced in recent years, with successful systems adopting denoising diffusion frameworks~\cite{ho2020denoising,sohl2015deep}. Early studies validated diffusion models for T2I and demonstrated advantages over GAN and autoregressive-based approaches~\cite{nichol2021glide,ramesh2022hierarchical,saharia2022photorealistic}. Latent diffusion—training the diffusion process in a compact latent space—proved especially effective at efficiency improvement and output resolution, and has become a \textit{de facto} standard in large-scale T2I systems (\textit{e.g.}, LDM and the Stable Diffusion family)~\cite{rombach2022high,esser2024scaling,podell2023sdxl}. Recent works replace the traditional UNet~\cite{ronneberger2015unet} backbone with transformer-based~\cite{attention_is_all_you_need} decoders, \textit{e.g.,} Diffusion Transformer (DiT) architectures~\cite{peebles2023scalable}, showing significant gains in image quality and scalability. These transformer decoders can better model long-range structure and complex compositions, which is central to our goal of conditioning high-fidelity generation on rich multimodal embeddings.


\noindent\textbf{Bridging large multimodal models and diffusion decoders.} Recently, integrating large language and multimodal models (LMMs/MLLMs) with diffusion decoders has attracted growing interest, enabling rich, structured, and interleaved text–image generation. Some approaches~\cite{li2024mini} translate complex multimodal instructions into textual or latent control codes that diffusion models can directly consume. Others introduce discrete or continuous visual tokenizers (\textit{e.g.}, Seed-Tokenizer~\cite{ge2307planting}, Seed-LLaMA~\cite{ge2023making}) that encode compact visual semantics to align language and vision token spaces for diffusion decoding. Jointly trained systems~\cite{jiang2026dreamvartamingreinforcedvisual} such as GILL~\cite{koh2023generating}, Emu~\cite{sun2023emu}, NExT-GPT~\cite{wu2024next}, and Any-GPT~\cite{zhan2024anygpt} further strengthen semantic alignment between multimodal embeddings and diffusion backbones. Methods like BLIP-Diffusion~\cite{li2023blip} extend this idea by projecting unified image–text representations into diffusion conditioning spaces to handle complex interleaved prompts. More recent pipelines—including UniFusion~\cite{li2025unifusion}, DreamEngine~\cite{chen2025multimodal}, Qwen-Image~\cite{wu2025qwen}, and EasyRef~\cite{zong2024easyref}—leverage pretrained MLLM or VLM features as conditioning signals for downstream diffusion transformers, enabling flexible text–image interleaving. However, these approaches typically rely only on the final-layer features of the MLLMs (\textit{e.g.,} Qwen-Image), or blend the ViT features from MLLMs with final-layer outputs via simple scalar mixing (\textit{e.g.,} DreamEngine). As a result, they often overlook fine-grained visual cues without relying on ID-relevant features (\textit{e.g.,} VAE), or provide only suboptimal identity preservation in subject-driven generation.

\section{Method}
\label{sec:method} 
\begin{figure*}[t!]
    \centering
    \includegraphics[width =\linewidth]{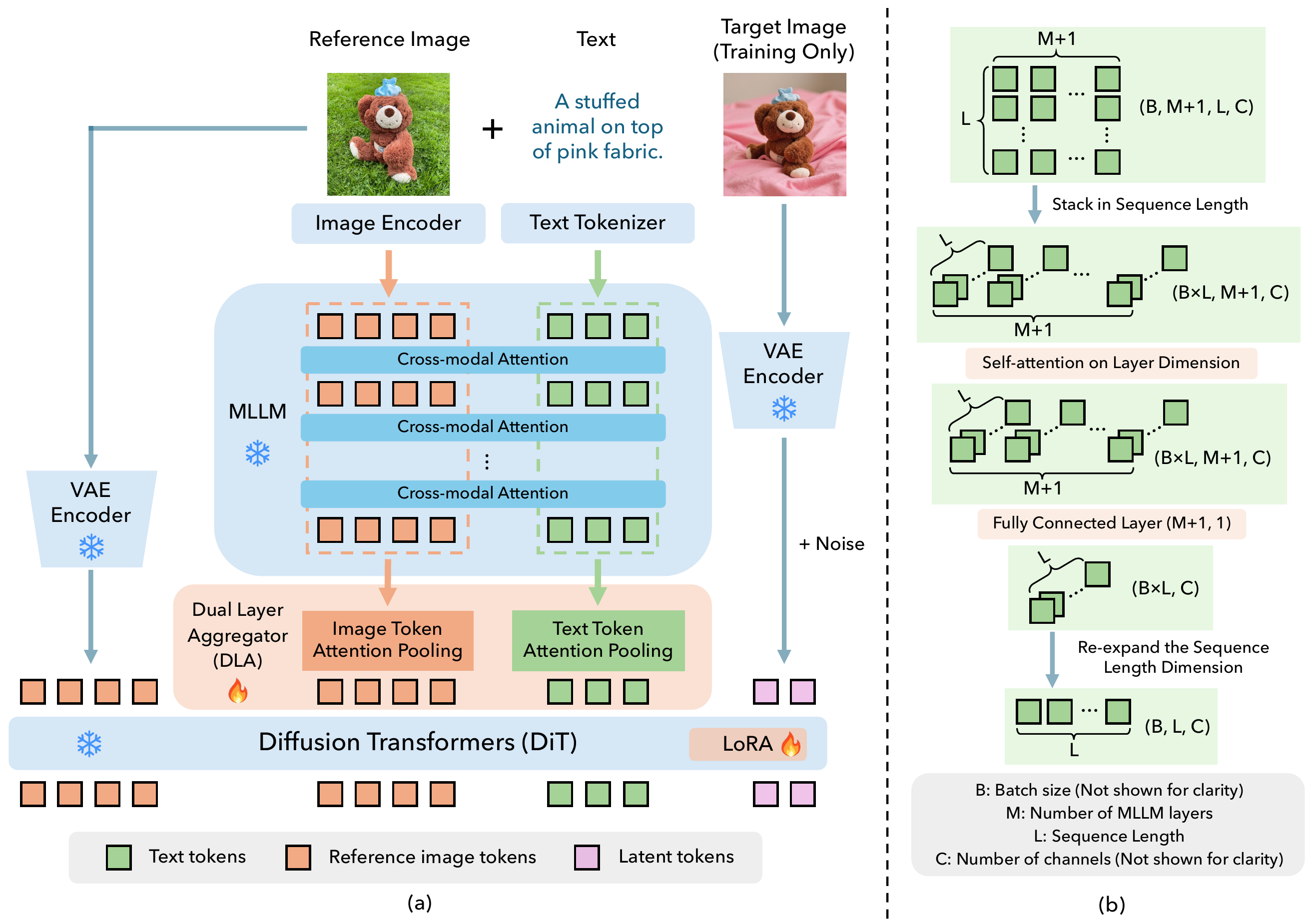}
    \caption{Overview of our framework. (a) The full model architecture, consisting of an MLLM for multimodal understanding, a VAE encoder for mapping images into latent space, a DiT backbone for diffusion denoising, and a Dual Layer Aggregator (DLA) that aligns MLLM embeddings for DiT. (b) Details of the token attention pooling module inside the DLA module, including its layerwise attention and pooling operations, to form MLLM aggregated embeddings from all $M$ MLLM layers and the input embedding layer.}
\vspace{-13pt}
    \label{fig:model}
\end{figure*}

Given a text prompt $\mathcal{T}$ and a set of reference images $\mathcal{I} = \{ I_n \}_{n=1}^{N}$, our method produces an image $\hat{I} = \mathcal{G}(\mathcal{T}, \mathcal{I})$ that aligns with the textual description while preserving the identity of the reference images. Our approach, visualized in Figure~\ref{fig:model}(a), is built on top of a Diffusion Transformer (DiT) backbone (Section~\ref{sec:prelim}) conditioned on a Multimodal Large Language Model (MLLM) and a VAE encoder. Specifically, we propose to use layerwise attention pooling (Section~\ref{sec:lap}) and propose a Dual Layer Aggregator (DLA) module (Section~\ref{sec:modality-specific}) that allows to extract aggregated features from MLLM layers for text and image modalities.
The architecture unifies MLLM for multimodal understanding and VAE for deriving high-fidelity identity details. To better reconcile capabilities of MLLM and VAE, we propose a multi-stage denoising process (Section~\ref{sec:denoising}) that allows integrating different conditioning branches and design a two-stage training strategy (Section~\ref{sec:two-stage-training}).


\subsection{Background: Diffusion Transformers}
\label{sec:prelim}

Diffusion models learn a mapping from a simple prior distribution to the data manifold through iterative denoising. Given a data sample $\mathbf{x}_0 \sim q(\mathbf{x}_0)$, the forward process gradually perturbs it with Gaussian noise under a variance schedule $\{\alpha_t\}_{t=1}^T$:
\begin{equation}
q(\mathbf{x}_t\mid\mathbf{x}_0) = \mathcal{N}(\mathbf{x}_t; \sqrt{\alpha_t}\mathbf{x}_0, (1-\alpha_t)\mathbf{I}),
\end{equation}
where $\mathbf{x}_T$ approaches an isotropic Gaussian. The reverse process is learned by predicting either the added noise or the clean sample $\mathbf{x}_0$ with the denoising network $\boldsymbol{\epsilon}_\theta(\mathbf{x}_t, t, \mathbf{c})$, conditioned on control signals $\mathbf{c}$ such as text or image embeddings. Recently, Rectified Flow~\cite{liu2022flow} reformulates diffusion as a deterministic \textit{transport process} parameterized by a time-dependent velocity field $\mathbf{v}_\theta(\mathbf{x}_t, t)$:
\begin{equation}
\frac{\mathrm{d}\mathbf{x}_t}{\mathrm{d}t} = \mathbf{v}_\theta(\mathbf{x}_t, t), \quad 
\mathbf{x}_0 = \mathbf{x}_1 + \int_0^1 \mathbf{v}_\theta(\mathbf{x}_t, t)\, \mathrm{d}t.
\end{equation}
This rectified formulation stabilizes training and simplifies inference by eliminating stochastic sampling steps. The objective becomes a velocity-matching loss:
\begin{equation}
\mathcal{L}_{\text{RF}} = 
\mathbb{E}_{t, \mathbf{x}_0, \mathbf{x}_1}
\big[\| \mathbf{v}_\theta(\mathbf{x}_t, t) - (\mathbf{x}_0 - \mathbf{x}_1) \|_2^2 \big].
\end{equation}

Building on this, Diffusion Transformers (DiT)~\cite{peebles2023scalable} replace the standard UNet backbone with a transformer that operates on \textit{patch tokens}. At each timestep $t$, the noisy image $\mathbf{x}_t$ is first projected into a latent representation $\mathbf{z}_t \in \mathbb{R}^{H \times W \times D}$. This latent is then flattened into a sequence of patch embeddings, augmented with timestep and conditioning tokens, and processed through self-attention layers to predict either the velocity or noise tokens for denoising.

In our experiments, we adopt \textit{FLUX.1 dev}~\cite{blackforestlabs2024flux}, a recent DiT-based architecture employing rectified flow parameterization as our backbone due to its training stability, synthesis capability, and modular conditioning design. Flux provides a flexible transformer-based diffusion decoder that seamlessly integrates multimodal embeddings, making it a strong foundation for our proposed MLLM-driven subject conditioning framework.

\subsection{Basic module: Layerwise attention pooling}\label{sec:lap}

Existing methods that connect MLLMs with diffusion models mainly focus on text-to-image generation and typically extract the single final layer feature as conditioning tokens~\cite{wu2025qwen, zong2024easyref}, assuming that the last layer contains the most informative semantic representation after multimodal reasoning. However, this strategy is suboptimal for subject-driven generation, where both \textit{text adherence} and \textit{identity preservation} are equally important.

\textbf{Motivation.} Since most MLLMs are optimized for high-level reasoning tasks such as VQA, their image tokens tend to lose fine-grained texture and appearance details in deeper layers. As also observed in~\cite{chen2024imageworth12tokens}, the visual representation in MLLMs shifts from low-level appearance to high-level semantics across layers when the layer dives deeper. This creates a \emph{representation mismatch}: no single layer provides both the semantic completeness required for text alignment and the fine-grained fidelity required for identity preservation. To alleviate this issue, we first explore to leverage a {Layerwise Attention Pooling (LAP)} mechanism inspired by UniFusion~\cite{li2025unifusion} that integrates features across multiple MLLM layers to retain both higher-level semantic and lower-level structural information.

\textbf{LAP Module.} Given MLLM feature maps from the input embedding and all transformer layers $\mathcal{F} = \{F_i\}_{i=0}^{M+1}$ ($M$ is the number of MLLM layers, and the additional one layer is the input embedding layer), where $F_i \in \mathbb{R}^{B \times L \times C}$ ($B$ is the batch size, $L$ is the sequence length, and $C$ is the channel number), LAP produces a summarized representation $\hat{F} \in \mathbb{R}^{B \times L \times C}$ via attention over the layer axis. Concretely, LAP implements a lightweight multi-head attention mechanism where the layer index is treated as the sequence dimension, followed by a fully connected projection for adaptive layer weighting, as shown in Figure~\ref{fig:model}(b).

\subsection{Dual layer aggregator}
\label{sec:modality-specific}
\begin{wrapfigure}[26]{r}{0.6\linewidth}
    \centering
    \includegraphics[width =\linewidth]
    {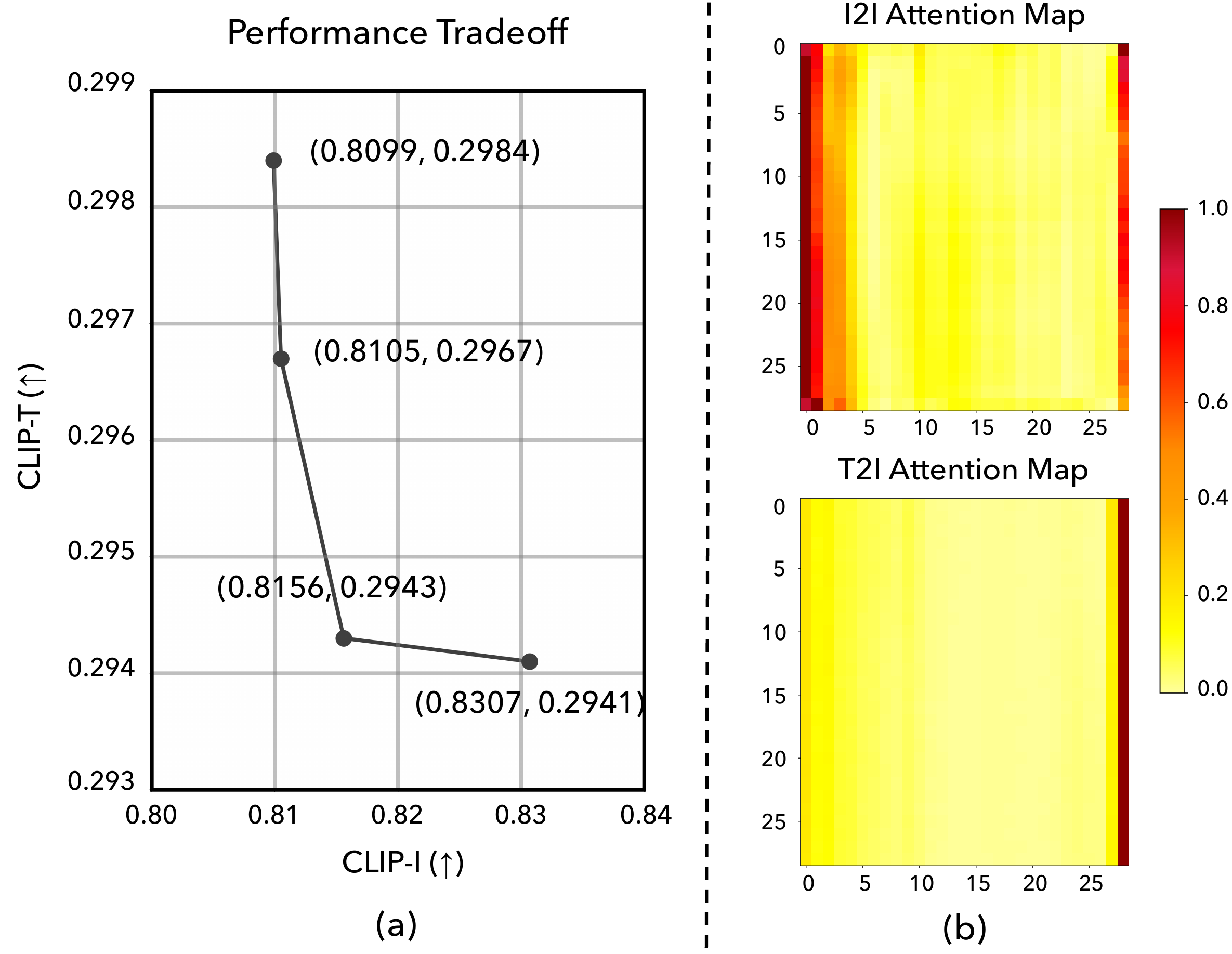}
    \vspace{-1.5mm}
    \caption{Preliminary analysis on single Layerwise Attention Pooling (LAP) across text and image modalities. (a) Performance tradeoff with different checkpoints when optimizing with a single LAP. (b) Attention maps from the model trained solely on I2I task and the model trained on T2I task, where both use single LAP model.}
    \label{fig:motivation}
\end{wrapfigure}

\textbf{Observation and motivation from the single LAP module.} As illustrated in Figure~\ref{fig:motivation}(a), preliminary experiments using a single LAP module to jointly summarize text and image tokens revealed a trade-off between identity preservation and text alignment for different checkpoints obtained during the optimization process. When trained together, the model tends to overfit to one modality, degrading the performance of the other. Further analysis in Figure~\ref{fig:motivation}(b) on text-to-image (T2I) and image-to-image (I2I) reconstruction tasks breaks down this issue, and shows that the layer-wise attention obtained from text and image tokens differ significantly, reflecting distinct hierarchical information patterns for each modality.

\textbf{DLA for multimodal processing.} Motivated by the observed issue, we introduce a {Dual Layer Aggregator (DLA)} that decouples layerwise aggregation across modalities. DLA consists of two separate LAP modules: one for text tokens and one for image tokens. The two modules have independent parameters, allowing them to learn different combinations of layers. Each LAP specializes in summarizing layerwise features most relevant to its modality—text LAP emphasizes on semantic fidelity and the prompt, while image LAP focuses on subject appearance and identity consistency. 
Importantly, this design does not sacrifice cross-modal interaction, as MLLMs already enable multimodal information to flow within intermediate layers, which means image tokens inside MLLM already absorb cross-modal information from text, and vice versa. Therefore, DLA avoids redundant multimodal fusion and instead focuses on modality-aware layerwise information processing. 

With the designed DLA module, each modality-specific LAP can focus on effectively aggregating intra-modal information without redundant fusion learning. Empirically, we observe that early and late layers in the MLLM often exhibit stronger activations corresponding to appearance and semantic cues, respectively. To maintain model-agnostic flexibility, we apply LAP to all MLLM layers, allowing DLA to adaptively learn each layer's contribution to identity or text following. This ensures robustness when adapting to different MLLM architectures with varying attention behaviors.


\subsection{Multi-stage timestep-aware denoising}
\label{sec:denoising}

The VAE encoder in diffusion models serves as a strong visual tokenizer that effectively captures detailed subject identity from reference images~\cite{wu2025less, cheng2025umo}. While VAEs preserve fine-grained appearance, they often suffer from copy-paste artifacts and lack semantic understanding. In contrast, MLLMs jointly encode text and images, offering better reasoning and layout understanding, but relatively weaker identity fidelity. To address the above limitations with single-source features, we leverage both conditioning sources to combine the complementary strengths of VAEs and MLLMs, and propose a \textit{multi-stage denoising process} that activates different conditioning branches along the denoising timesteps. This design aligns with the inherent coarse-to-fine nature of diffusion: earlier steps capture semantics and global layout, and later steps refine local details. Specifically, MLLM conditioning is used in early steps for semantic and compositional reasoning; both MLLM and VAE conditioning are applied in the middle for balanced control; and only VAE conditioning is used in late steps for detailed identity refinement.

\noindent
\textbf{Formulation.}
The denoising network predicts the clean sample at each step as:
\begin{equation}
    \mathbf{\hat{x}}_{t-1} = f_{\theta}(\mathbf{x_t}, \mathbf{c}_{\text{MLLM}} \cdot M_{\text{MLLM}}(t), \mathbf{c}_{\text{VAE}} \cdot M_{\text{VAE}}(t)),
\end{equation}
where $f_{\theta}$ denotes the denoising transformer, and $\mathbf{c}_{\text{MLLM}}$ and $\mathbf{c}_{\text{VAE}}$ are conditioning embeddings from the two encoders. The timestep-dependent masks $M_{\text{MLLM}}(t), M_{\text{VAE}}(t) \in \{0, 1\}$ control which branches are active. During training, the reference image input for either branch (MLLM or VAE) is randomly dropped to ensure robustness. As a result, the whole system can naturally handle scenarios when only one of the sources has the reference image input.


We define three denoising stages parameterized by $\tau_1$ and $\tau_2$:
\begin{equation}
M_{\text{MLLM}}(t), M_{\text{VAE}}(t) =
\begin{cases}
(1, 0), & t \geq \tau_1 \quad \text{(early)} \\
(1, 1), & \tau_2 \leq t < \tau_1 \quad \text{(middle)} \\
(0, 1), & t < \tau_2 \quad \text{(late)}.
\end{cases}
\end{equation}
\textbf{Integration with rectified flow.}
This stage-aware conditioning naturally integrates with our rectified flow objective. As the rectified flow continuously transports samples from noise to data, the conditioning signal shifts from semantic alignment via the MLLM, to fine-detailed identity refinement via the VAE near the data manifold, achieving coherent and instruction-aware subject generation.

\subsection{Two-stage training strategy}
\label{sec:two-stage-training}

Training a diffusion system conditioned on both MLLM and VAE embeddings presents a unique challenge. 
Since our timestep-aware denoising process requires the model to function when only one of the two modalities (MLLM or VAE) is present, both encoders must independently learn to contribute meaningful signals for subject-driven generation. 
To achieve this, we adopt the following \textit{two-stage training strategy}.

In the first stage, we train the diffusion transformer using only MLLM-derived conditioning. 
This stage encourages the MLLM to fully exploit its multimodal reasoning ability, and capture identity-related cues from the reference images as well. 
In the second stage, we jointly train the entire framework—MLLM, VAE, and DiT—enabling the model to balance high-level reasoning from the MLLM with fine-grained identity features from the VAE.

This staged optimization prevents the VAE from dominating identity preservation too early. 
If trained jointly from scratch, the VAE tends to absorb most of the identity learning, leaving the MLLM under-optimized and ineffective in the early denoising steps—where global structure and appearance are primarily determined. 
Consequently, once identity information is far off track in the early timesteps, it cannot be recovered later even when VAE conditioning is introduced. 
Our two-stage strategy therefore ensures both conditioning pathways to contribute effectively throughout the denoising process, leading to harmonized identity fidelity and prompt alignment.

\section{Experiments}
\label{sec:exp}

\begin{figure*}[t]
    \centering
    \includegraphics[width =\linewidth]{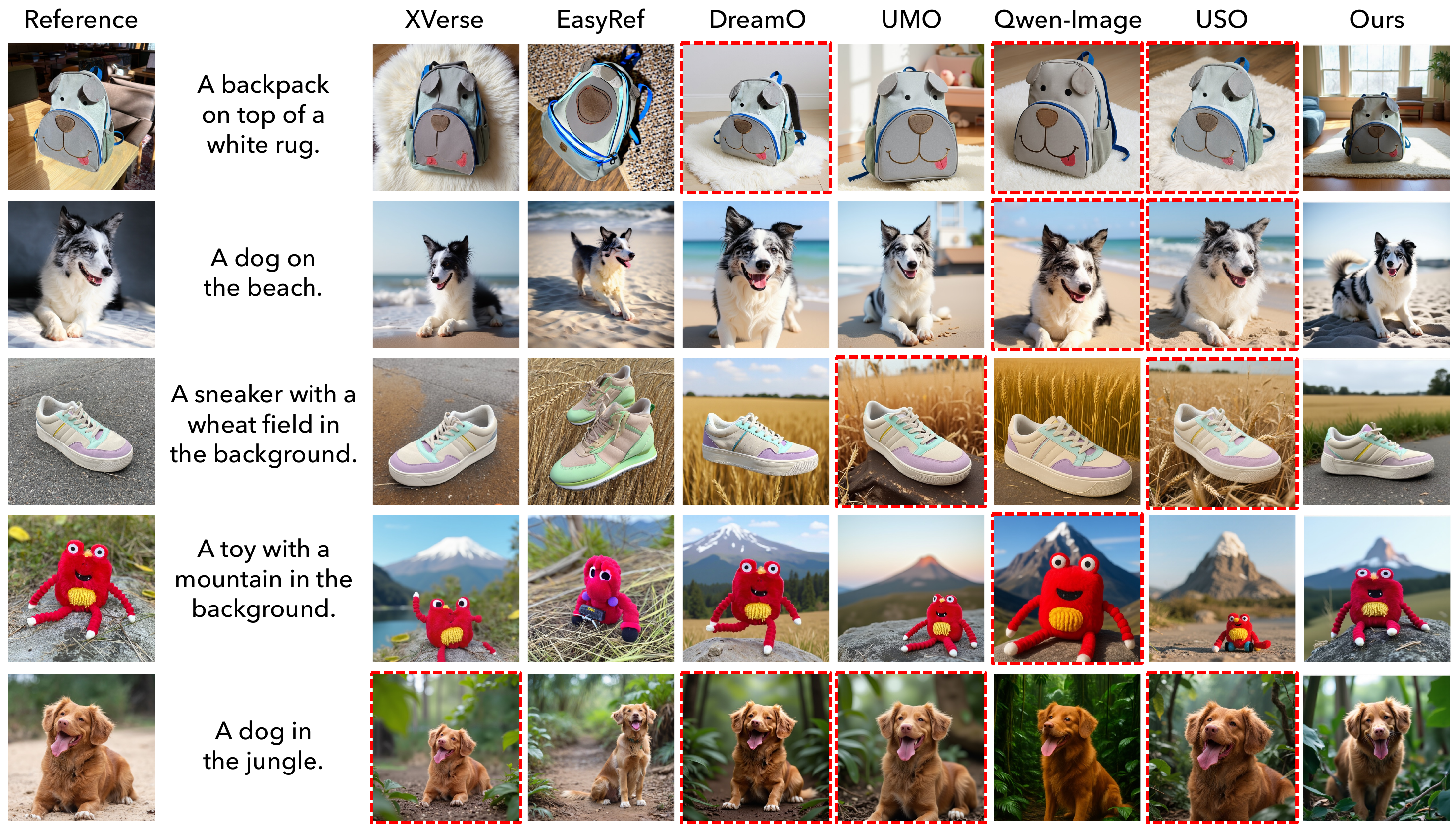}
    \caption{Comparisons of our method with state-of-the-art subject-driven generation approaches. \textbf{\textit{\textcolor{red}{Red dashed lines}}} indicate instances where other methods suffer from the copy-paste issue. In contrast, our method produces images that maintain strong subject identity, exhibit creative pose variations, and respect underlying physical constraints.}
    \label{fig:main_qual}
\end{figure*}
%
%

\subsection{Experimental settings}

\noindent\textbf{Dataset.}  
To explore the potential of MLLMs for subject-driven generation, we use only public datasets throughout our experiments. Our model is trained on the publicly available UNO-1M~\cite{wu2025less}, which contains approximately 400K image pairs after filtering with MLLM-based scoring criteria. Each pair features a subject with matched images of the same identity.

\noindent\textbf{Implementation details.}  
Following the two-stage training strategy described in Section~\ref{sec:two-stage-training}, we first train the MLLM-DiT framework for 25K steps, and then incorporate both MLLM and VAE as conditioning signals for an additional 10K steps. Training is conducted on 8 NVIDIA H100 GPUs, each with a batch size of 16, using a constant learning rate of $1\text{e-}5$. 

We adopt InternVL3-8B~\cite{zhu2025internvl3} as the MLLM and FLUX.1 dev~\cite{blackforestlabs2024flux} as the DiT, with a LoRA rank of 512 for finetuning the DiT attention blocks. The MLLM and other weights in DiT are all frozen. During inference, we set timestep-aware denoising thresholds to $\tau_1=0.95$ and $\tau_2=0.85$, use a cosine denoising schedule, and apply a classifier-free guidance (CFG) value of 2.5 for all stages when evaluating metrics. Section~\ref{sec:ablation} further analyzes how these parameters affect performance and provide users with finer control over identity fidelity, pose variation, and overall image quality.

\subsection{Comparisons with existing methods}
In this section, we conduct comprehensive experiments on various aspects to demonstrate the capability of our method. Besides the \textbf{(1)} standard benchmark evaluations, \textbf{(2)} we propose an evaluation criteria to quantify the copy-paste issue and illustrate that the issue gets largely mitigated. Also, \textbf{(3)} better multimodal understanding capability is revealed with both qualitative and quantitative results. Additionally, \textbf{(4)} automatic human-aligned evaluation and \textbf{(5)} user study demonstrate that our method receives more preference from users compared to existing models.

\noindent\textbf{(1) Standard benchmark performance.} 
We compare our model with state-of-the-art subject-driven generation methods including OminiControl~\cite{tan2024ominicontrol}, OmniGen2~\cite{wu2025omnigen2}, UNO~\cite{wu2025less}, XVerse~\cite{chen2025xverse}, DreamO~\cite{mou2025dreamo}, USO~\cite{wu2025uso}, and UMO~\cite{cheng2025umo}, as well as recent approaches that connect MLLMs with diffusion models, including DreamEngine~\cite{chen2025multimodal}, Qwen-Image~\cite{wu2025qwen}, and EasyRef~\cite{zong2024easyref}. 

\begin{wraptable}[19]{r}{0.52\linewidth}
    \centering
    \caption{Quantitative comparison on DreamBench. $^\dag$Training with the same public data as ours (single-subject data from UNO-1M). First block indicates VAE-based methods while the second block indicates MLLM-based approaches.}
    \rowcolors{2}{white}{uoftcoolgray!25}
    \centering
    \resizebox{\linewidth}{!}{
        \begin{tabular}{lccc}
        \toprule
          \textbf{Method}   &  \textbf{DINO-I ($\uparrow$)}  & \textbf{CLIP-I ($\uparrow$)} & \textbf{CLIP-T ($\uparrow$)} \\ 
        \midrule
        OminiControl~\cite{tan2024ominicontrol} & 0.5987 & 0.7840 & 0.3186 \\ 
        OminiControl$^\dag$~\cite{tan2024ominicontrol} & 0.6537 & 0.7979 & 0.3019 \\ 
        OmniGen2~\cite{wu2025omnigen2} & 0.7323 & 0.8268 & 0.3185\\ 
        UNO~\cite{wu2025less} & 0.7484 & 0.8354 & 0.3040 \\ 
        UNO$^\dag$~\cite{wu2025less} & 0.6860 & 0.8161  & 0.3071\\ 
        XVerse~\cite{chen2025xverse} & 0.7215 & 0.8175& 0.3098\\ 
        DreamO~\cite{mou2025dreamo} & \textbf{0.7537} & 0.8356& 0.3086\\ 
        USO~\cite{wu2025uso} & 0.7478 & 0.8263 & \textbf{0.3213}\\ 
        UMO~\cite{cheng2025umo} & 0.7481 & 0.8339& 0.3022\\ 
        \midrule
        DreamEngine~\cite{chen2025multimodal}  &  0.5195& 0.7428& 0.3006\\ 
        Qwen-Image~\cite{wu2025qwen}  & 0.7317 & 0.8261 & 0.3158\\
        EasyRef~\cite{zong2024easyref} & 0.6961 & 0.8153 & 0.3031\\ 
          Ours (MLLM only) & 0.6788 & 0.8228  & 0.2988\\ 
          Ours (MLLM + VAE) & 0.7482 & \textbf{0.8443} & 0.3010\\ 
        \bottomrule
        \end{tabular}
    }
    \vspace{-10pt}
    \label{tab:main_quantitative}
\end{wraptable}
As many existing systems rely on private high-quality subject-driven datasets, we also re-train UNO which has public training code with the same UNO-1M data to better show the potential of our method. Following previous works, DreamBench~\cite{ruiz2023dreambooth} is adopted as our main evaluation benchmark for experimental analysis and ablation study. DINO-I~\cite{caron2021emergingpropertiesselfsupervisedvision} and CLIP-I~\cite{radford2021learningtransferablevisualmodels} are used for measuring identity similarity, and CLIP-T is used for text-image alignment. As shown in Table~\ref{tab:main_quantitative}, our MLLM-only model (only trained for the first stage with the MLLM-DiT framework) already reaches performance on par with UNO trained under the same conditions, demonstrating the strength of our DLA in extracting multimodal features and identity signals from MLLMs. With both MLLM and VAE conditioning, our full model—trained entirely on public data—achieves performance comparable to state-of-the-art methods. Qualitative comparisons in Figure~\ref{fig:main_qual} show that our approach produces \textit{more diverse poses} while preserving identity, and yields \textit{more physically coherent} scenes, avoiding artifacts such as subjects floating above backgrounds. Beyond the standard DreamBench, we also include evaluations on additional benchmarks including XVerseBench~\cite{chen2025xverse} and a multi-subject LAMICBench~\cite{chen2025lamic} on our model with slight multi-subject adaptation in the Appendix.

\begin{wraptable}[13]{r}{0.52\linewidth}
    \vspace{-12pt}
    \centering
    \caption{Quantitative comparisons of subject variation between the reference and generated images, measuring the copy-paste issue. We evaluate differences in azimuth and polar angles to assess the subject pose diversity produced by the model, showing its ability to generate more varied poses and reduce copy-paste artifacts.}
\rowcolors{2}{white}{uoftcoolgray!25}
    \centering
    \resizebox{\linewidth}{!}{
        \begin{tabular}{lccc}
        \toprule
          \textbf{Method}   &  \textbf{Azimuth ($\uparrow$)}   & \textbf{Polar ($\uparrow$)} & \textbf{Average ``Recall'' Rate ($\downarrow$)}\\ 
        \midrule
        OmniGen2~\cite{wu2025omnigen2} & 22.6 & 7.0  &0.486\\
        DreamO~\cite{mou2025dreamo} &  22.1 & 9.6  & 0.372\\
        USO~\cite{wu2025uso}   &  20.8 & 9.6  & 0.401 \\ 
        Qwen-Image~\cite{wu2025qwen} & 17.6 &  7.8  & 0.460\\ 
        \midrule
        Ours & \textbf{25.7} & \textbf{10.4}  & \textbf{0.349}  \\ 
        \bottomrule
        \end{tabular}
    }
    \vspace{-10pt}
    \label{tab:copy_paste_issue}
\end{wraptable}
\noindent\textbf{(2) Copy-paste issue alleviation.} A common failure mode in subject-driven generation—particularly in VAE-based methods—is the \textit{copy-paste} effect, where the generated subject closely mimics the reference pose with minimal variation. This issue is largely overlooked in the prior evaluations in existing works that mainly focus on identity preservation and text alignment. As illustrated in Figure~\ref{fig:main_qual}, many existing approaches suffer from this behavior (highlighted with red dashed boxes), whereas our method produces subjects with noticeably more diverse and creative poses. To quantify this effect, we adopt Orient Anything~\cite{wang2025orient} to estimate the azimuth and polar angles of subjects in both the reference and generated images, and compute their average orientation discrepancy. We further propose a \textit{``Recall''@\,$k^{\circ}$} metric—the percentage of generated samples whose orientation angles (both azimuth and polar) are below $k^{\circ}$, and report the \textit{Average ``Recall'' Rate} metric, which is averaged over $k^{\circ}\in\{5^{\circ}, 10^{\circ}, 15^{\circ}, 20^{\circ}\}$ for \textit{``Recall''@\,$k^{\circ}$}. As shown in Table~\ref{tab:copy_paste_issue}, our MLLM-based conditioning significantly mitigates the copy-paste issue compared to previous methods. While the diversity of the generated subject can also be reflected in other factors (\textit{e.g.}, posture), orientation is a crucial and easily measurable indicator, especially for rigid objects, so it serves as a practical proxy for evaluating the copy-paste artifacts.

\begin{figure}[t]
\centering

\begin{minipage}[t]{0.48\textwidth}
\vspace{-0pt}
    \centering
    \includegraphics[width =\linewidth]{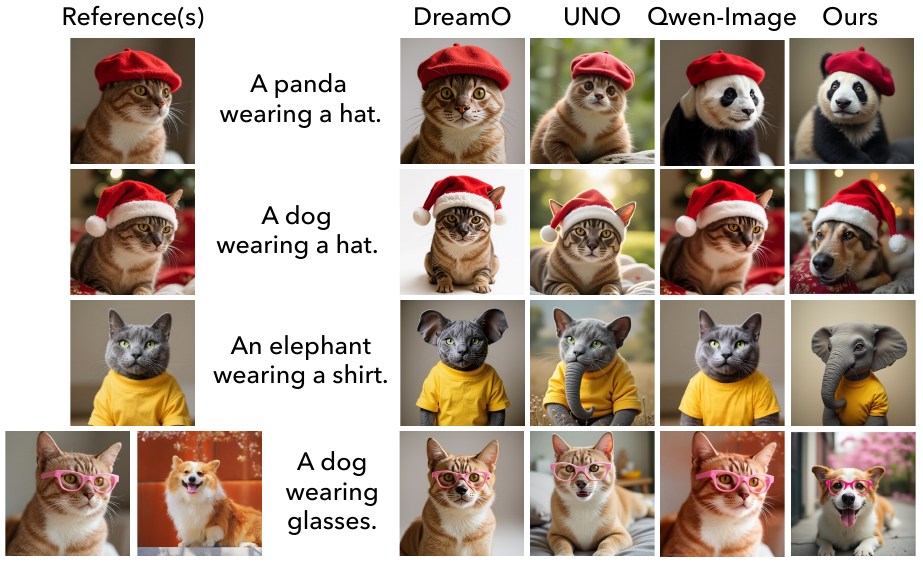}
    \caption{Comparisons on reasoning capability show that VAE-based methods often fail on complex user prompts, producing copy-pasted subjects or incorrect concept binding. MLLM-DiT pipelines like Qwen-Image also struggle in understanding these challenging user prompts, demonstrating the prior solution for connecting MLLM and DiT is suboptimal. In contrast, our method, conditioned solely on MLLM signals, accurately interprets the prompts and associates concepts with the appropriate visual elements.}
\vspace{-12pt}
    \label{fig:reasoning_qual}
\end{minipage}
\hfill
\begin{minipage}[t]{0.48\textwidth}
\vspace{5pt}
    \rowcolors{2}{white}{uoftcoolgray!25}
    \centering
    \captionof{table}{Quantitative comparisons on the constructed benchmark with 350 samples for multimodal reasoning capability in subject-driven generation.}
    \vspace{-6pt}
    \resizebox{0.8\linewidth}{!}{
        \begin{tabular}{lcccc}
        \toprule
        Metric & UNO & DreamO & Qwen-Image & Ours \\ 
        \midrule
        CLIP-T ($\uparrow$) & 0.2851 & 0.2888 & 0.3099 &\textbf{0.3208} \\ 
        \bottomrule
        \end{tabular}
    }
\label{tab:mm_reasoning_subject_driven}
    
    \vspace{0.2cm}

\centering
\vspace{-2pt}
\captionof{table}{Quantitative comparisons on the human-aligned MLLM-based scores on the subject categories in DreamBench++.}
\vspace{2pt}
\rowcolors{0}{}{}
\resizebox{\textwidth}{!}{
\begin{tabular}{lcccccccc}
\toprule
\multirow{2}{*}{Method} &
\multicolumn{8}{c}{\textbf{MLLM-based Scores (0-4 scale) ($\uparrow$)}}  \\
\cmidrule(lr){2-9}
 & (A) & (B) & (C) & (D) & (E) & (F) &  (G) & Average \\
\midrule
\rowcolor{uoftcoolgray!25}
DreamO & 2.837 & 2.892 & 2.802 & 3.402 & 2.737 & 2.462 & 2.737 & 2.838 \\
UNO & 2.539 & 2.753 & 2.474 & 3.027 & 2.303 & 2.103 & 2.576 & 2.539 \\
\rowcolor{uoftcoolgray!25}
USO & 2.790 & 2.868 & 2.798 & 3.410 & 2.663 & 2.400 & 2.668 &  2.800\\
Ours & \textbf{3.119} & \textbf{2.969} & \textbf{3.006} & \textbf{3.568} & \textbf{2.962} & \textbf{2.601} & \textbf{2.847} & \textbf{3.010} \\
\bottomrule
\end{tabular}%
}
\label{tab:vlm_based_comparison}
    

    \captionof{table}{User study from 30 participants with a total of 1,500 votes on samples from DreamBench and XVerseBench.}
    \vspace{2pt}
    \rowcolors{2}{white}{uoftcoolgray!25}
    \centering
    \resizebox{0.95\linewidth}{!}{
        \begin{tabular}{lccccc}
        \toprule
        Method & XVerse & DreamO & USO & UMO & Ours \\ 
        \midrule
        Score (1-10 scale) ($\uparrow$) & 5.75 & 6.31 & 6.74 & 6.02 & \textbf{7.26}\\ 
        \bottomrule
        \end{tabular}
    }
    \vspace{-10pt}
    \label{tab:user_study}
    
\end{minipage}
\vspace{-6pt}
\end{figure}

\noindent\textbf{(3) Reasoning capability.} 
The text prompts in DreamBench are relatively simple and require limited cross-modal reasoning, making differences in text-following performance across models small. Methods like USO can achieve high scores despite occasionally \textit{``copy-pasting''} the subject and placing it on top of the prompted background. More challenging scenarios arise when the user prompt refers to concepts that are not the sole focus of the reference image, or when concept binding needs to be figured out between the text and the visual input. Figure~\ref{fig:reasoning_qual} illustrates such cases, where correct generation depends on understanding and reasoning over both modalities. For instance, in the first row, the model must associate the ``hat'' mentioned in the prompt with the correct region of the reference image, while ignoring irrelevant distractors such as the cat. VAE-based pipelines struggle here because they encode text and reference images independently, limiting their ability to jointly interpret user intent. Qwen-Image with the MLLM-DiT structure also shows multiple failure cases, suggesting that conditioning the DiT solely on the final layer features does not fully leverage the multimodal reasoning capacity of MLLMs. In contrast, our model successfully aligns text and image cues, producing coherent outputs. Notably, even though our model is \textit{trained only on single-subject data}, it can take two reference images as input (last row of Figure~\ref{fig:reasoning_qual}) and still correctly bind textual concepts to the appropriate visual regions. To further quantitatively verify the multimodal reasoning capability, we construct a benchmark consisting of 350 samples similar to the examples in Figure~\ref{fig:reasoning_qual}, and evaluate on the text following capability that largely reflects the correctness of the generated images from user instructions. Detailed information about the constructed benchmark can be referred in the supplementary material. As shown in Table~\ref{tab:mm_reasoning_subject_driven}, our method greatly outperforms the existing models on multimodal understanding on these complex scenarios, because the MLLM in our model jointly encode both images and text that enables cross-modal concept binding and reasoning.
\textbf{(4) Human-aligned evaluation.} To provide an additional perspective regarding human-aligned preference, we perform evaluation on DreamBench++~\cite{peng2025dreambench}, which adopts an MLLM-based scoring metric that is aligned with human perception. MLLMs are expected to give an overall score from 0-4 on subject consistency following prompts used in~\cite{peng2025dreambench, zheng2025track, kumar2025deft} for the generated images, considering multiple aspects including shape, color, and texture. More details about the prompts used for MLLMs are described in the supplementary material. We select seven MLLMs with different architectures and sizes to foster the soundness of the evaluation: (A) GPT-4o~\cite{gpt4o} (original choice from DreamBench++), (B) Gemma 3 27B~\cite{gemmateam2025gemma3technicalreport}, (C) Gemini 2.5 Flash~\cite{comanici2025gemini25pushingfrontier}, (D) Gemini 3 Flash~\cite{gemini_30_flash}, (E) Qwen3-VL-30B-A3B-Thinking~\cite{bai2025qwen3vltechnicalreport}, (F) Qwen3-VL-235B-A22B-Thinking~\cite{bai2025qwen3vltechnicalreport}, and (G) Mistral-Small-3.2-24B-Instruct~\cite{mistral_small}. Table~\ref{tab:vlm_based_comparison} demonstrate the superiority on subject consistency of our method evaluated with all types of MLLMs.

\textbf{(5) User study.} To further calibrate subject-driven generation quality with human perception, we conduct user study on the images generated by XVerse, DreamO, USO, UMO, and our method. We randomly select 10 samples from DreamBench and XVerseBench, and ask the volunteers to score the generated assets in a scale of 1-10 on the overall quality, where the participants are guided to focus on subject fidelity, text following, \textit{etc}. that are considered important factors for subject-driven generation. Details on the instructions and interface of our user study can be referred in the supplementary material. There are 30 volunteers participating in our user study, and a total of 1,500 votes are collected. The results of the user study in Table~\ref{tab:user_study} show that our method is also more subjectively preferred from real user experience.

\subsection{Ablation study}
\label{sec:ablation}
In this section, we provide the ablation study and analysis on the design and training mechanisms of our method. Due to the limit of space, more ablation study and analysis can be found in the Appendix.

\begin{wraptable}[18]{r}{0.6\linewidth}
    \centering
    \caption{Analysis on different strategies of connecting MLLM features to the DiT. The first block reports baselines that rely on last-layer conditioning and their variants. The second block evaluates single LAP configurations, showing that a single LAP for both text and image tokens preserves identity reasonably well, but severely weakens text following.}
\vspace{-4pt}
    \rowcolors{2}{white}{uoftcoolgray!25}
    \centering
    \resizebox{\linewidth}{!}{
        \begin{tabular}{lccccc}
        \toprule
        \textbf{Method} & 
        \makecell[c]{\textbf{Selected}\\\textbf{Layers}} & 
        \makecell[c]{\textbf{Residual}\\\textbf{Connection}} & 
        \textbf{DINO-I ($\uparrow$)} & 
        \textbf{CLIP-I ($\uparrow$)} & 
        \textbf{CLIP-T ($\uparrow$)} \\ 
        \midrule
        Last layer \cite{wu2025qwen} & - & - & 0.6566 & 0.8128 & 0.2893\\ 
        Last layer (blend ViT)~\cite{chen2025multimodal}   & - & - & 0.7118 & 0.8286 & 0.2850\\ 
        Last layer (mix ViT)      & - & - & 0.7097 & 0.8233 & 0.2946\\ 
        \midrule
        
        Single LAP & 0-9 &$\times$ & 0.7167 & 0.8391 & 0.2969\\ 
        Single LAP    & 10-19 & $\times$ & 0.7315 & 0.8463 & 0.2957\\ 
        Single LAP    & 20-28 & $\times$ & 0.7325 & 0.8386 & 0.2981\\ 
        Single LAP    & 0-28  & $\times$ & 0.7524 & 0.8502 & 0.2878\\ 
        Single LAP    & 0-9   & $\checkmark$ & 0.7282 & 0.8497 & 0.2944\\ 
        Single LAP    & 10-19 & $\checkmark$ & 0.7109 & 0.8377 & 0.2958\\ 
         Single LAP   & 20-28 & $\checkmark$ & 0.7246 & 0.8389 & 0.2963\\ 
         Single LAP   & 0-28  & $\checkmark$ & 0.7242 & 0.8379 & 0.2974\\ 
        \midrule
        
        DLA (Dual LAP)  & 0-28 & $\checkmark$ & 0.7275 & 0.8401 & 0.3013\\ 
        DLA (Dual LAP)  & 0-28 & $\times$ & 0.7482 & 0.8443 & 0.3010\\ 
        \bottomrule
        \end{tabular}
    }
    \label{tab:ab1_strategy}
    \vspace{-5pt}
\end{wraptable}
\noindent\textbf{(1) Strategies for leveraging MLLM.}  
We conduct a systematic analysis to identify effective ways to squeeze MLLM capacity to diffusion models for subject-driven generation. We ablate several strategies, including those from prior works: last-layer feature extraction from Qwen-Image~\cite{wu2025qwen}, scalar blending of MLLM ViT feature with last-layer feature in DreamEngine~\cite{chen2025multimodal} (blend ViT), and concatenation of MLLM ViT image feature with last-layer text feature (mix ViT). Additionally, we explore different layer selections in our LAP module by partitioning InternVL3-8B’s 28 layers into early (0–9), middle (10–19), and late (20–28) layers, and evaluate residual connections to the last layer. Table~\ref{tab:ab1_strategy} shows that existing strategies are suboptimal under limited data and resource constraints. Using a single LAP for both text and image preserves identity reasonably but severely compromises text alignment. Residual connections to the last layer do not improve performance and can even degrade it, suggesting that overemphasizing last-layer features may be harmful.

\begin{figure}[t]
\centering

\begin{minipage}[t]{0.47\textwidth}
\vspace{-0pt}
    \centering
    \includegraphics[width =0.9\linewidth]{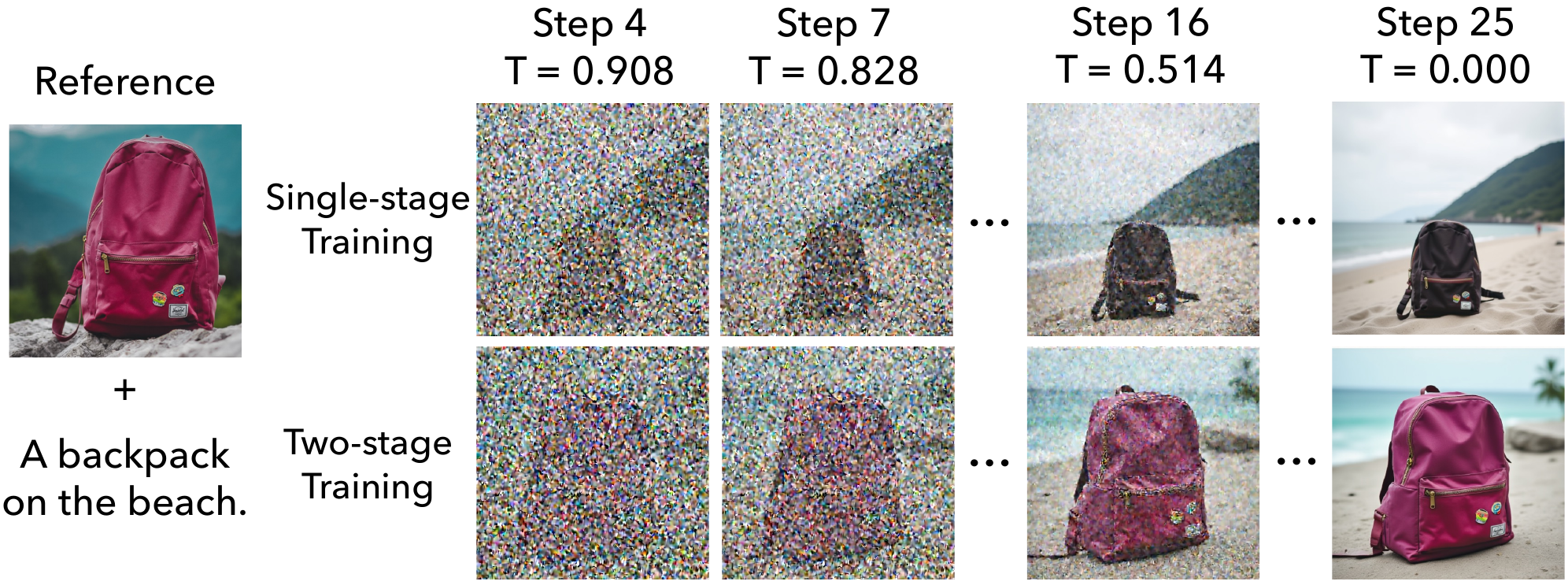}
    \vspace{-2mm}
    \caption{Single-stage training prevents the model from leveraging timestep-aware denoising, limiting both potential performance gains and the flexibility for user control.}
\vspace{-10pt}
    \label{fig:ab_single_stage_training_no_msd}
\end{minipage}
\hfill
\begin{minipage}[t]{0.49\textwidth}
\captionof{table}{Comparison regarding single-stage training, with and without timestep-aware denoising (TAD). The results highlight the importance of our two-stage training strategy, which first establishes a well-trained MLLM-DiT system before introducing the VAE.}
\vspace{-4pt}
\rowcolors{2}{white}{uoftcoolgray!25}
    \centering
    \resizebox{\linewidth}{!}{
        \begin{tabular}{lccc}
        \toprule
          \textbf{Method}   &  \textbf{DINO-I ($\uparrow$)}  & \textbf{CLIP-I ($\uparrow$)} & \textbf{CLIP-T ($\uparrow$)} \\ 
        \midrule

        Single-stage Training w/o TAD  & 0.7184 & 0.8245& 0.2971\\ 
        Single-stage Training with TAD  & 0.5763 & 0.7686& 0.2995\\ 
        \midrule
        Two-stage Training & \textbf{0.7482}  & \textbf{0.8443} & \textbf{0.3010}\\ 
        \bottomrule
        \end{tabular}
    }
    \label{tab:single_stage}
    \vspace{-15pt}
\end{minipage}
\vspace{-5pt}
\end{figure}

\noindent\textbf{(2) Two-stage training.} 
Optimizing a framework that integrates both MLLM and VAE for subject-driven generation is non-trivial. In Section~\ref{sec:two-stage-training}, we propose a two-stage training strategy: first training the MLLM-DiT framework in the initial stage and then adding the VAE in the second stage. Without this staged approach, the capacity from MLLMs for identity preservation cannot be sufficiently unleashed, which prevents the model from fully leveraging the timestep-aware denoising process. As shown in Figure~\ref{fig:ab_single_stage_training_no_msd}, the initial denoising steps conditioned on MLLM largely determine the final appearance of the image; if MLLM has not developed decent identity preservation capability, the VAE in the later stages cannot correct the denoising trajectory. Table~\ref{tab:single_stage} further shows that the model trained with single-stage strategy has inferior performance in both identity preservation and text alignment, and its failure to utilize timestep-aware denoising to boost the performance. This occurs because VAE tokens, which are originally optimized for reconstruction, dominate the generation process, therefore reducing the information contribution from MLLM features. As a result, the DiT loses one source of identity information, and the cross-modal reasoning and understanding capabilities from the MLLM are diminished as well.

\vspace{-4pt}
\section{Conclusion}
\label{sec:conclusion}




We study towards the optimal strategy to utilize MLLMs for subject-driven generation, with the introduced Dual Layer Aggregation (DLA) module. Our analysis shows that aggregating representations across all layers and aligning text and visual modalities separately, is critical to achieving strong multimodal understanding and identity preservation. Combined with the VAE's strength in capturing fine-grained visual details, our multi-stage denoising framework and two-stage training strategy further harmonize the conditioning signals from both MLLM and VAE, and provide users with more flexibility during generation. 






{
    \small
    \bibliographystyle{ieeenat_fullname}
\bibliography{bib/2d_editing,bib/3d_editing,bib/4d_editing,bib/baselines,bib/dynamic,bib/related}
}

\newpage
\clearpage
\section*{NeurIPS Paper Checklist}

\begin{enumerate}

\item {\bf Claims}
    \item[] Question: Do the main claims made in the abstract and introduction accurately reflect the paper's contributions and scope?
    \item[] Answer: \answerYes{} 
    \item[] Justification: The claims made in the abstract and introduction accurately reflect the paper’s
contributions and scope.
    \item[] Guidelines:
    \begin{itemize}
        \item The answer \answerNA{} means that the abstract and introduction do not include the claims made in the paper.
        \item The abstract and/or introduction should clearly state the claims made, including the contributions made in the paper and important assumptions and limitations. A \answerNo{} or \answerNA{} answer to this question will not be perceived well by the reviewers. 
        \item The claims made should match theoretical and experimental results, and reflect how much the results can be expected to generalize to other settings. 
        \item It is fine to include aspirational goals as motivation as long as it is clear that these goals are not attained by the paper. 
    \end{itemize}

\item {\bf Limitations}
    \item[] Question: Does the paper discuss the limitations of the work performed by the authors?
    \item[] Answer: \answerYes{} 
    \item[] Justification: The limitations of the work have been discussed in App.~\ref{sec:discussions} in the Appendix.
    \item[] Guidelines:
    \begin{itemize}
        \item The answer \answerNA{} means that the paper has no limitation while the answer \answerNo{} means that the paper has limitations, but those are not discussed in the paper. 
        \item The authors are encouraged to create a separate ``Limitations'' section in their paper.
        \item The paper should point out any strong assumptions and how robust the results are to violations of these assumptions (e.g., independence assumptions, noiseless settings, model well-specification, asymptotic approximations only holding locally). The authors should reflect on how these assumptions might be violated in practice and what the implications would be.
        \item The authors should reflect on the scope of the claims made, e.g., if the approach was only tested on a few datasets or with a few runs. In general, empirical results often depend on implicit assumptions, which should be articulated.
        \item The authors should reflect on the factors that influence the performance of the approach. For example, a facial recognition algorithm may perform poorly when image resolution is low or images are taken in low lighting. Or a speech-to-text system might not be used reliably to provide closed captions for online lectures because it fails to handle technical jargon.
        \item The authors should discuss the computational efficiency of the proposed algorithms and how they scale with dataset size.
        \item If applicable, the authors should discuss possible limitations of their approach to address problems of privacy and fairness.
        \item While the authors might fear that complete honesty about limitations might be used by reviewers as grounds for rejection, a worse outcome might be that reviewers discover limitations that aren't acknowledged in the paper. The authors should use their best judgment and recognize that individual actions in favor of transparency play an important role in developing norms that preserve the integrity of the community. Reviewers will be specifically instructed to not penalize honesty concerning limitations.
    \end{itemize}

\item {\bf Theory assumptions and proofs}
    \item[] Question: For each theoretical result, does the paper provide the full set of assumptions and a complete (and correct) proof?
    \item[] Answer: \answerNA{} 
    \item[] Justification: The paper does not include theoretical results.
    \item[] Guidelines:
    \begin{itemize}
        \item The answer \answerNA{} means that the paper does not include theoretical results. 
        \item All the theorems, formulas, and proofs in the paper should be numbered and cross-referenced.
        \item All assumptions should be clearly stated or referenced in the statement of any theorems.
        \item The proofs can either appear in the main paper or the supplemental material, but if they appear in the supplemental material, the authors are encouraged to provide a short proof sketch to provide intuition. 
        \item Inversely, any informal proof provided in the core of the paper should be complemented by formal proofs provided in appendix or supplemental material.
        \item Theorems and Lemmas that the proof relies upon should be properly referenced. 
    \end{itemize}

    \item {\bf Experimental result reproducibility}
    \item[] Question: Does the paper fully disclose all the information needed to reproduce the main experimental results of the paper to the extent that it affects the main claims and/or conclusions of the paper (regardless of whether the code and data are provided or not)?
    \item[] Answer: \answerYes{} 
    \item[] Justification: We have discussed all the details to reproduce all experimental results in our
paper.
    \item[] Guidelines:
    \begin{itemize}
        \item The answer \answerNA{} means that the paper does not include experiments.
        \item If the paper includes experiments, a \answerNo{} answer to this question will not be perceived well by the reviewers: Making the paper reproducible is important, regardless of whether the code and data are provided or not.
        \item If the contribution is a dataset and\slash or model, the authors should describe the steps taken to make their results reproducible or verifiable. 
        \item Depending on the contribution, reproducibility can be accomplished in various ways. For example, if the contribution is a novel architecture, describing the architecture fully might suffice, or if the contribution is a specific model and empirical evaluation, it may be necessary to either make it possible for others to replicate the model with the same dataset, or provide access to the model. In general. releasing code and data is often one good way to accomplish this, but reproducibility can also be provided via detailed instructions for how to replicate the results, access to a hosted model (e.g., in the case of a large language model), releasing of a model checkpoint, or other means that are appropriate to the research performed.
        \item While NeurIPS does not require releasing code, the conference does require all submissions to provide some reasonable avenue for reproducibility, which may depend on the nature of the contribution. For example
        \begin{enumerate}
            \item If the contribution is primarily a new algorithm, the paper should make it clear how to reproduce that algorithm.
            \item If the contribution is primarily a new model architecture, the paper should describe the architecture clearly and fully.
            \item If the contribution is a new model (e.g., a large language model), then there should either be a way to access this model for reproducing the results or a way to reproduce the model (e.g., with an open-source dataset or instructions for how to construct the dataset).
            \item We recognize that reproducibility may be tricky in some cases, in which case authors are welcome to describe the particular way they provide for reproducibility. In the case of closed-source models, it may be that access to the model is limited in some way (e.g., to registered users), but it should be possible for other researchers to have some path to reproducing or verifying the results.
        \end{enumerate}
    \end{itemize}

\item {\bf Open access to data and code}
    \item[] Question: Does the paper provide open access to the data and code, with sufficient instructions to faithfully reproduce the main experimental results, as described in supplemental material?
    \item[] Answer: \answerNo{} 
    \item[] Justification:  While the current submission does not contain code implementation, we have
included detailed instructions in the Appendix to provide guidance for reproducing the
experimental results. Also, we promise that we will open-source the data and code after
paper acceptance.
    \item[] Guidelines:
    \begin{itemize}
        \item The answer \answerNA{} means that paper does not include experiments requiring code.
        \item Please see the NeurIPS code and data submission guidelines (\url{https://neurips.cc/public/guides/CodeSubmissionPolicy}) for more details.
        \item While we encourage the release of code and data, we understand that this might not be possible, so \answerNo{} is an acceptable answer. Papers cannot be rejected simply for not including code, unless this is central to the contribution (e.g., for a new open-source benchmark).
        \item The instructions should contain the exact command and environment needed to run to reproduce the results. See the NeurIPS code and data submission guidelines (\url{https://neurips.cc/public/guides/CodeSubmissionPolicy}) for more details.
        \item The authors should provide instructions on data access and preparation, including how to access the raw data, preprocessed data, intermediate data, and generated data, etc.
        \item The authors should provide scripts to reproduce all experimental results for the new proposed method and baselines. If only a subset of experiments are reproducible, they should state which ones are omitted from the script and why.
        \item At submission time, to preserve anonymity, the authors should release anonymized versions (if applicable).
        \item Providing as much information as possible in supplemental material (appended to the paper) is recommended, but including URLs to data and code is permitted.
    \end{itemize}

\item {\bf Experimental setting/details}
    \item[] Question: Does the paper specify all the training and test details (e.g., data splits, hyperparameters, how they were chosen, type of optimizer) necessary to understand the results?
    \item[] Answer: \answerYes{} 
    \item[] Justification: The training and test details have been specified in the paper, which are enough to understand the results.
    \item[] Guidelines:
    \begin{itemize}
        \item The answer \answerNA{} means that the paper does not include experiments.
        \item The experimental setting should be presented in the core of the paper to a level of detail that is necessary to appreciate the results and make sense of them.
        \item The full details can be provided either with the code, in appendix, or as supplemental material.
    \end{itemize}

\item {\bf Experiment statistical significance}
    \item[] Question: Does the paper report error bars suitably and correctly defined or other appropriate information about the statistical significance of the experiments?
    \item[] Answer: \answerNo{} 
    \item[] Justification: Retraining the model is computationally demanding, so we do not run multiple times for reporting the statistical significance.
    \item[] Guidelines:
    \begin{itemize}
        \item The answer \answerNA{} means that the paper does not include experiments.
        \item The authors should answer \answerYes{} if the results are accompanied by error bars, confidence intervals, or statistical significance tests, at least for the experiments that support the main claims of the paper.
        \item The factors of variability that the error bars are capturing should be clearly stated (for example, train/test split, initialization, random drawing of some parameter, or overall run with given experimental conditions).
        \item The method for calculating the error bars should be explained (closed form formula, call to a library function, bootstrap, etc.)
        \item The assumptions made should be given (e.g., Normally distributed errors).
        \item It should be clear whether the error bar is the standard deviation or the standard error of the mean.
        \item It is OK to report 1-sigma error bars, but one should state it. The authors should preferably report a 2-sigma error bar than state that they have a 96\% CI, if the hypothesis of Normality of errors is not verified.
        \item For asymmetric distributions, the authors should be careful not to show in tables or figures symmetric error bars that would yield results that are out of range (e.g., negative error rates).
        \item If error bars are reported in tables or plots, the authors should explain in the text how they were calculated and reference the corresponding figures or tables in the text.
    \end{itemize}

\item {\bf Experiments compute resources}
    \item[] Question: For each experiment, does the paper provide sufficient information on the computer resources (type of compute workers, memory, time of execution) needed to reproduce the experiments?
    \item[] Answer: \answerYes{} 
    \item[] Justification: The compute resources required for reproduce the is specified in Section~\ref{sec:exp} in the Appendix.
    \item[] Guidelines:
    \begin{itemize}
        \item The answer \answerNA{} means that the paper does not include experiments.
        \item The paper should indicate the type of compute workers CPU or GPU, internal cluster, or cloud provider, including relevant memory and storage.
        \item The paper should provide the amount of compute required for each of the individual experimental runs as well as estimate the total compute. 
        \item The paper should disclose whether the full research project required more compute than the experiments reported in the paper (e.g., preliminary or failed experiments that didn't make it into the paper). 
    \end{itemize}
    
\item {\bf Code of ethics}
    \item[] Question: Does the research conducted in the paper conform, in every respect, with the NeurIPS Code of Ethics \url{https://neurips.cc/public/EthicsGuidelines}?
    \item[] Answer: \answerYes{} 
    \item[] Justification: The conducted research conforms with the NeurIPS Code of Ethics in every respect.
    \item[] Guidelines:
    \begin{itemize}
        \item The answer \answerNA{} means that the authors have not reviewed the NeurIPS Code of Ethics.
        \item If the authors answer \answerNo, they should explain the special circumstances that require a deviation from the Code of Ethics.
        \item The authors should make sure to preserve anonymity (e.g., if there is a special consideration due to laws or regulations in their jurisdiction).
    \end{itemize}

\item {\bf Broader impacts}
    \item[] Question: Does the paper discuss both potential positive societal impacts and negative societal impacts of the work performed?
    \item[] Answer: \answerYes{} 
    \item[] Justification:  Both positive and negative societal impacts have been discussed in App.~\ref{sec:societal_impact} in the Appendix.
    \item[] Guidelines:
    \begin{itemize}
        \item The answer \answerNA{} means that there is no societal impact of the work performed.
        \item If the authors answer \answerNA{} or \answerNo, they should explain why their work has no societal impact or why the paper does not address societal impact.
        \item Examples of negative societal impacts include potential malicious or unintended uses (e.g., disinformation, generating fake profiles, surveillance), fairness considerations (e.g., deployment of technologies that could make decisions that unfairly impact specific groups), privacy considerations, and security considerations.
        \item The conference expects that many papers will be foundational research and not tied to particular applications, let alone deployments. However, if there is a direct path to any negative applications, the authors should point it out. For example, it is legitimate to point out that an improvement in the quality of generative models could be used to generate Deepfakes for disinformation. On the other hand, it is not needed to point out that a generic algorithm for optimizing neural networks could enable people to train models that generate Deepfakes faster.
        \item The authors should consider possible harms that could arise when the technology is being used as intended and functioning correctly, harms that could arise when the technology is being used as intended but gives incorrect results, and harms following from (intentional or unintentional) misuse of the technology.
        \item If there are negative societal impacts, the authors could also discuss possible mitigation strategies (e.g., gated release of models, providing defenses in addition to attacks, mechanisms for monitoring misuse, mechanisms to monitor how a system learns from feedback over time, improving the efficiency and accessibility of ML).
    \end{itemize}
    
\item {\bf Safeguards}
    \item[] Question: Does the paper describe safeguards that have been put in place for responsible release of data or models that have a high risk for misuse (e.g., pre-trained language models, image generators, or scraped datasets)?
    \item[] Answer: \answerNA{} 
    \item[] Justification: Our paper does not have such risks.
    \item[] Guidelines:
    \begin{itemize}
        \item The answer \answerNA{} means that the paper poses no such risks.
        \item Released models that have a high risk for misuse or dual-use should be released with necessary safeguards to allow for controlled use of the model, for example by requiring that users adhere to usage guidelines or restrictions to access the model or implementing safety filters. 
        \item Datasets that have been scraped from the Internet could pose safety risks. The authors should describe how they avoided releasing unsafe images.
        \item We recognize that providing effective safeguards is challenging, and many papers do not require this, but we encourage authors to take this into account and make a best faith effort.
    \end{itemize}

\item {\bf Licenses for existing assets}
    \item[] Question: Are the creators or original owners of assets (e.g., code, data, models), used in the paper, properly credited and are the license and terms of use explicitly mentioned and properly respected?
    \item[] Answer: \answerYes{} 
    \item[] Justification: We have properly credited and listed the license in App.~\ref{sec:licenses} in the Appendix.
    \item[] Guidelines:
    \begin{itemize}
        \item The answer \answerNA{} means that the paper does not use existing assets.
        \item The authors should cite the original paper that produced the code package or dataset.
        \item The authors should state which version of the asset is used and, if possible, include a URL.
        \item The name of the license (e.g., CC-BY 4.0) should be included for each asset.
        \item For scraped data from a particular source (e.g., website), the copyright and terms of service of that source should be provided.
        \item If assets are released, the license, copyright information, and terms of use in the package should be provided. For popular datasets, \url{paperswithcode.com/datasets} has curated licenses for some datasets. Their licensing guide can help determine the license of a dataset.
        \item For existing datasets that are re-packaged, both the original license and the license of the derived asset (if it has changed) should be provided.
        \item If this information is not available online, the authors are encouraged to reach out to the asset's creators.
    \end{itemize}

\item {\bf New assets}
    \item[] Question: Are new assets introduced in the paper well documented and is the documentation provided alongside the assets?
    \item[] Answer: \answerNA{} 
    \item[] Justification: At the submission time, we have not released the new assets used in the paper.
We will well document our data and model at the time we release our data after acceptance.
    \item[] Guidelines:
    \begin{itemize}
        \item The answer \answerNA{} means that the paper does not release new assets.
        \item Researchers should communicate the details of the dataset\slash code\slash model as part of their submissions via structured templates. This includes details about training, license, limitations, etc. 
        \item The paper should discuss whether and how consent was obtained from people whose asset is used.
        \item At submission time, remember to anonymize your assets (if applicable). You can either create an anonymized URL or include an anonymized zip file.
    \end{itemize}

\item {\bf Crowdsourcing and research with human subjects}
    \item[] Question: For crowdsourcing experiments and research with human subjects, does the paper include the full text of instructions given to participants and screenshots, if applicable, as well as details about compensation (if any)? 
    \item[] Answer: \answerYes{} 
    \item[] Justification: We have included the full text of instructions given to the participants in
App.~\ref{sec:user_study} in the appendix, as well as the screenshots of the user study.
    \item[] Guidelines:
    \begin{itemize}
        \item The answer \answerNA{} means that the paper does not involve crowdsourcing nor research with human subjects.
        \item Including this information in the supplemental material is fine, but if the main contribution of the paper involves human subjects, then as much detail as possible should be included in the main paper. 
        \item According to the NeurIPS Code of Ethics, workers involved in data collection, curation, or other labor should be paid at least the minimum wage in the country of the data collector. 
    \end{itemize}

\item {\bf Institutional review board (IRB) approvals or equivalent for research with human subjects}
    \item[] Question: Does the paper describe potential risks incurred by study participants, whether such risks were disclosed to the subjects, and whether Institutional Review Board (IRB) approvals (or an equivalent approval/review based on the requirements of your country or institution) were obtained?
    \item[] Answer: \answerNA{} 
    \item[] Justification: There is no potential risks in the crowdsourcing research conducted for this work.
    \item[] Guidelines:
    \begin{itemize}
        \item The answer \answerNA{} means that the paper does not involve crowdsourcing nor research with human subjects.
        \item Depending on the country in which research is conducted, IRB approval (or equivalent) may be required for any human subjects research. If you obtained IRB approval, you should clearly state this in the paper. 
        \item We recognize that the procedures for this may vary significantly between institutions and locations, and we expect authors to adhere to the NeurIPS Code of Ethics and the guidelines for their institution. 
        \item For initial submissions, do not include any information that would break anonymity (if applicable), such as the institution conducting the review.
    \end{itemize}

\item {\bf Declaration of LLM usage}
    \item[] Question: Does the paper describe the usage of LLMs if it is an important, original, or non-standard component of the core methods in this research? Note that if the LLM is used only for writing, editing, or formatting purposes and does \emph{not} impact the core methodology, scientific rigor, or originality of the research, declaration is not required.
    \item[] Answer: \answerNA{} 
    \item[] Justification: The core method development in this research does not involve LLMs as any
important, original, or non-standard components.
    \item[] Guidelines:
    \begin{itemize}
        \item The answer \answerNA{} means that the core method development in this research does not involve LLMs as any important, original, or non-standard components.
        \item Please refer to our LLM policy in the NeurIPS handbook for what should or should not be described.
    \end{itemize}

\end{enumerate}


\newpage

\appendix

\renewcommand\thesection{\Alph{section}}
\renewcommand{\thefigure}{\Alph{figure}}
\renewcommand{\thetable}{\Alph{table}}
\renewcommand\thealgorithm{\Alph{algorithm}}

\setcounter{section}{0}
\setcounter{equation}{0}
\setcounter{figure}{0}
\setcounter{table}{0}
\setcounter{algorithm}{0}

{\Large\centering\bf 	
Squeezing Capacity from Multimodal Large Language Models for Subject-driven Generation\\}

\vspace{4mm}
{\large\centering\rmfamily Technical Appendices and Supplementary Material\\}

\vspace{4mm}

In this appendix, we provide additional analyses and experiments to further validate the effectiveness of our method along with more discussions. First, in Section~\ref{sec:layer_analysis}, we conduct a layer-wise analysis of the DLA module during inference to understand its contribution across different layers. Section~\ref{sec:layer_selection} explores various layer selection strategies for training DLA, analyzing both efficiency and performance trade-offs. 
Section~\ref{sec:sensitivity} conducts the sensitivity analysis to demonstrate the robustness of hyperparameter choices and flexible user control from our method.
In Section~\ref{sec:mlmm_ablation}, we perform an ablation study on different MLLM backbones, demonstrating the robustness and generalizability of our approach. We further show the adaptability of our method to multi-subject generation in Section~\ref{sec:multi_subject} with small-scale finetuning. More details about the construction of the multimodal reasoning benchmark, additional implementation details for reproducibility, unsuccessful attempts behind the scenes, the instruction and interface used for the user study, and the prompts used for MLLM-based evaluation are explained in Sections~\ref{sec:mm_reasoning_construction},~\ref{sec:additional_implementation},~\ref{sec:unsuccessful_attempts},~\ref{sec:user_study},~\ref{sec:vlm_based_evaluation}, respectively. More quantitative and qualitative evaluations are displayed in Sections~\ref{sec:more_benchmarks} and~\ref{sec:qualitative} to highlight the visual advantages of our approach. Section~\ref{sec:licenses} include the license and terms of use for the models and data used in the paper. Finally, Sections~\ref{sec:discussions} and~\ref{sec:societal_impact} present the limitations, discussions, and potential societal impact of our method.

\section{Layer analysis for DLA at inference time}
\label{sec:layer_analysis}
In this section, we analyze how different layers in the dual text LAP and image LAP of our DLA contribute to performance. Specifically, we take the fully trained model and selectively zero out certain layers during inference to examine their impact. The results in Table~\ref{tab:ab_sup_inference_diff_layers} reveal three key observations. First, the image modality is highly sensitive to the removal of the early MLLM layers, suggesting that these layers are essential for preserving fine-grained visual details. Second, the text modality shows more robustness to layer removal. This indicates that, although the text modality primarily relies on the later layers, the model can still retrieve similar textual information from the preceding layers when later layers are removed. Third, we find that disabling one modality can occasionally bring slight improvements when the other modality is partially dropped, implying that the model may rely more heavily on a single modality in such cases. This further supports our claim that balancing the two modalities is crucial for optimal performance.

We further include qualitative comparisons in Figure~\ref{fig:supp_figure_inference_comparison_1} and Figure~\ref{fig:supp_figure_inference_comparison_2}, which visually support the observations drawn from Table~\ref{tab:ab_sup_inference_diff_layers}. Specifically, we observe two consistent trends.
First, when early layers of the image modality are dropped (e.g., zeroing out layers 0-19), the model struggles to preserve identity, whereas using only early layers achieves comparable ID consistency. Second, for the text modality, the later layers appear more critical as using only layers 0-9 significantly weakens the model’s ability to understand and follow the prompt.

\begin{table}[h]
    \rowcolors{2}{white}{uoftcoolgray!25}
    \centering
    \caption{Layer analysis of our DLA during inference. We assess the contribution of each MLLM layer in our full DLA by selectively zeroing out text and image modalities during inference. For simplicity, the numerical results of difference are rounded to two digits after the decimal point.}
    \vspace{5pt}
    \resizebox{0.8\linewidth}{!}{
        \begin{tabular}{ccccc}
        \toprule
        \textbf{Image Layers} & 
        \textbf{Text Layers} & 
         
        \textbf{DINO-I ($\uparrow$)} & 
        \textbf{CLIP-I ($\uparrow$)} & 
        \textbf{CLIP-T ($\uparrow$)} \\ 
        \midrule
        0-28 & 0-28  & 0.7482 & 0.8443 & 0.3010 \\ 
        \midrule
        \textcolor{gray}{\sout{0-9}} 10-19 20-28 & 0-28  & 0.6368 {\color{red} (-0.11)} & 0.7959  {\color{red} (-0.05)}  & 0.3111 {\color{blue} (+0.01)}\\ 
        0-9 \textcolor{gray}{\sout{10-19}} 20-28 & 0-28 & 0.7472 {\color{red} (-0.00)} & 0.8439 {\color{red} (-0.00)} & 0.3011 {\color{blue} (+0.00)}\\
        0-9 10-19 \textcolor{gray}{\sout{20-28}} & 0-28 & 0.7344 {\color{red} (-0.01)} & 0.8353 {\color{red} (-0.01)} & 0.3041 {\color{blue} (+0.00)}\\
        \textcolor{gray}{\sout{0-9}} 10-19 \textcolor{gray}{\sout{20-28}} & 0-28 & 0.6058 {\color{red} (-0.14)} & 0.7837 {\color{red} (-0.06)} & 0.3117 {\color{blue} (+0.01)}\\
        0-9 \textcolor{gray}{\sout{10-19}} \textcolor{gray}{\sout{20-28}} & 0-28& 0.7093 {\color{red} (-0.04)} & 0.8251 {\color{red} (-0.02)} & 0.3067 {\color{blue} (+0.01)}\\
        \textcolor{gray}{\sout{0-9}} \textcolor{gray}{\sout{10-19}} 20-28 & 0-28& 0.5898 {\color{red} (-0.16)} & 0.7773 {\color{red} (-0.07)} & 0.3129 {\color{blue} (+0.01)}\\
        \textcolor{gray}{\sout{0-12}} 13-16 \textcolor{gray}{\sout{17-28}} & 0-28& 0.5962 {\color{red} (-0.15)} & 0.7769 {\color{red} (-0.07)} & 0.3134 {\color{blue} (+0.01)}\\
        \textcolor{gray}{\sout{0-24}} 25-28 & 0-28& 0.5560 {\color{red} (-0.19)} & 0.7618 {\color{red} (-0.08)} & 0.3129 {\color{blue} (+0.01)}\\
        0-3 \textcolor{gray}{\sout{4-28}} & 0-28& 0.5493 {\color{red} (-0.20)} & 0.7609 {\color{red} (-0.08)} & 0.3126 {\color{blue} (+0.01)}\\
        \midrule
        0-28 & \textcolor{gray}{\sout{0-9}} 10-19 20-28& 0.7557 {\color{blue} (+0.01)} & 0.8474 {\color{blue} (+0.00)} & 0.3006 {\color{red} (-0.00)}\\
        0-28 & 0-9 \textcolor{gray}{\sout{10-19}} 20-28& 0.7399 {\color{red} (-0.01)} & 0.8405 {\color{red} (-0.00)} & 0.2990 {\color{red} (-0.00)}\\

        0-28 & 0-9 10-19 \textcolor{gray}{\sout{20-28}}& 0.7267 {\color{red} (-0.02)} & 0.8354 {\color{red} (-0.01)} & 0.2991 {\color{red} (-0.00)}\\
        
        0-28 & \textcolor{gray}{\sout{0-9}} \textcolor{gray}{\sout{10-19}} 20-28& 0.7560 {\color{blue} (+0.01)} & 0.8498 {\color{blue} (+0.01)} & 0.2966 {\color{red} (-0.00)}\\
        
        0-28 & \textcolor{gray}{\sout{0-9}} 10-19 \textcolor{gray}{\sout{20-28}} & 0.7363 {\color{red} (-0.01)} & 0.8402 {\color{red} (-0.00)} & 0.3018 {\color{blue} (+0.00)}\\
        0-28 & 0-9 \textcolor{gray}{\sout{10-19}} \textcolor{gray}{\sout{20-28}} & 0.7473 {\color{red} (-0.00)} & 0.8492 {\color{blue} (+0.00)} & 0.2545 {\color{red} (-0.05)}\\
        0-28 & \textcolor{gray}{\sout{0-12}} 13-16 \textcolor{gray}{\sout{17-28}} & 0.7661 {\color{blue} (+0.02)} & 0.8631 {\color{blue} (+0.02)} & 0.2657 {\color{red} (-0.04)}\\
        0-28 & \textcolor{gray}{\sout{0-24}} 25-28 & 0.8274 {\color{blue} (+0.08)} & 0.9068 {\color{blue} (+0.06)} & 0.2480 {\color{red} (-0.05)}\\
        0-28 & 0-3 \textcolor{gray}{\sout{4-28}} & 0.7823 {\color{blue} (+0.03)} & 0.8724 {\color{blue} (+0.03)} & 0.2590 {\color{red} (-0.04)}\\
        \midrule
        0-9 \textcolor{gray}{\sout{10-19}} \textcolor{gray}{\sout{20-28}} & 0-9 \textcolor{gray}{\sout{10-19}} \textcolor{gray}{\sout{20-28}}& 0.6950 {\color{red} (-0.05)} & 0.8153 {\color{red} (-0.03)} & 0.2586 {\color{red} (-0.04)}\\
        0-9 \textcolor{gray}{\sout{10-19}} \textcolor{gray}{\sout{20-28}} & \textcolor{gray}{\sout{0-9}} 10-19 \textcolor{gray}{\sout{20-28}}& 0.6906 {\color{red} (-0.06)} & 0.8173 {\color{red} (-0.03)} & 0.3080 {\color{blue} (+0.01)}\\
        0-9 \textcolor{gray}{\sout{10-19}} \textcolor{gray}{\sout{20-28}} & \textcolor{gray}{\sout{0-9}} \textcolor{gray}{\sout{10-19}} 20-28& 0.7135 {\color{red} (-0.03)} & 0.8301 {\color{red} (-0.01)} & 0.3042 {\color{blue} (+0.00)}\\
        \textcolor{gray}{\sout{0-9}} 10-19 \textcolor{gray}{\sout{20-28}} & 0-9 \textcolor{gray}{\sout{10-19}} \textcolor{gray}{\sout{20-28}}& 0.4852 {\color{red} (-0.26)} & 0.7129 {\color{red} (-0.13)} & 0.2582 {\color{red} (-0.04)}\\
        \textcolor{gray}{\sout{0-9}} 10-19 \textcolor{gray}{\sout{20-28}} & \textcolor{gray}{\sout{0-9}} 10-19 \textcolor{gray}{\sout{20-28}} & 0.5743 {\color{red} (-0.17)} & 0.7700 {\color{red} (-0.07)} & 0.3135 {\color{blue} (+0.01)}\\
        \textcolor{gray}{\sout{0-9}} 10-19 \textcolor{gray}{\sout{20-28}} & \textcolor{gray}{\sout{0-9}} \textcolor{gray}{\sout{10-19}} 20-28& 0.6139 {\color{red} (-0.13)} & 0.7878 {\color{red} (-0.06)} & 0.3085 {\color{blue} (+0.01)}\\
        \textcolor{gray}{\sout{0-9}} \textcolor{gray}{\sout{10-19}} 20-28 & 0-9 \textcolor{gray}{\sout{10-19}} \textcolor{gray}{\sout{20-28}}& 0.4144 {\color{red} (-0.33)} & 0.6820 {\color{red} (-0.16)} & 0.2500 {\color{red} (-0.05)}\\
        \textcolor{gray}{\sout{0-9}} \textcolor{gray}{\sout{10-19}} 20-28 & \textcolor{gray}{\sout{0-9}} 10-19 \textcolor{gray}{\sout{20-28}}& 0.5562 {\color{red} (-0.19)} & 0.7644 {\color{red} (-0.08)} & 0.3139 {\color{blue} (+0.01)}\\
        \textcolor{gray}{\sout{0-9}} \textcolor{gray}{\sout{10-19}} 20-28 & \textcolor{gray}{\sout{0-9}} \textcolor{gray}{\sout{10-19}} 20-28& 0.6038 {\color{red} (-0.14)} & 0.7848 {\color{red} (-0.06)} & 0.3099 {\color{blue} (+0.01)}\\
        \bottomrule
        \end{tabular}
    }
    \label{tab:ab_sup_inference_diff_layers}
\end{table}

\section{Layer selection for training DLA}
\label{sec:layer_selection}
In the main paper, we primarily ablate layer selection strategies for the single LAP. Here, we extend the investigation with a more comprehensive analysis of dual LAP layer strategies within our DLA module. It is important to note the key difference between this experiment and the previous one in Table~\ref{tab:ab_sup_inference_diff_layers}. In Section~\ref{sec:layer_analysis}, the study was conducted during inference using a fully trained model where individual layers were selectively zeroed out. In contrast, the experiments presented here involve training the entire pipeline from scratch while using only a subset of MLLM layers for either the text LAP or the image LAP.

Thus, the analysis in Section~\ref{sec:layer_analysis} emphasizes the contribution of each individual group of layers within DLA, whereas this section focuses on evaluating different strategies for connecting MLLM features to DiT. The results are shown in Table~\ref{tab:ab_sup_select_double_lap_layers}, and our key observations are summarized as follows. First, pre-selecting early layers (0-9) yields noticeable performance gains for identity metrics, likely because the model leans more toward copy-paste behavior, as text-following performance drops.
Second, almost all layer-preselection strategies for the text modality lead to degraded performance, aligning with the finding from Section~\ref{sec:layer_analysis} that textual information is distributed across all MLLM layers. Third, despite using only subset layers, the diffusion model can still attain comparable or even improved performance for both modalities, suggesting that the current layer aggregation design may not fully exploit the representational efficiency of different layers. Some layers may be redundant while others are more informative, potentially depending on the specific context.

\begin{table}[t]
    \rowcolors{2}{white}{uoftcoolgray!25}
    \centering
    \caption{Layer selection of our DLA during training. We re-train each of our variants of DLA by selecting different parts of layers for text and image modalities.}
    \vspace{5pt}
    \resizebox{0.85\linewidth}{!}{
        \begin{tabular}{ccccc}
        \toprule
        \makecell[c]{\textbf{Selected Image}\\\textbf{LAP Layers}}  & 
        \makecell[c]{\textbf{Selected Text}\\\textbf{LAP Layers}}  & 
         
        \textbf{DINO-I ($\uparrow$)} & 
        \textbf{CLIP-I ($\uparrow$)} & 
        \textbf{CLIP-T ($\uparrow$)} \\ 
        \midrule
        0-28 & 0-28 & 0.7482 & 0.8443 & 0.3010\\
        \midrule
        0-9 & 0-28 & 0.7781 {\color{blue} (+0.03)} & 0.8567 {\color{blue} (+0.01)} & 0.2932 {\color{red} (-0.01)}\\
        10-19 & 0-28 & 0.7519 {\color{blue} (+0.00)} & 0.8424 {\color{red} (-0.00)} & 0.2972 {\color{red} (-0.00)}\\
        20-28 & 0-28 & 0.7189 {\color{red} (-0.03)} & 0.8292 {\color{red} (-0.02)} & 0.2990 {\color{red} (-0.00)}\\
        \midrule
        0-28 & 0-9 & 0.7464 {\color{red} (-0.00)} & 0.8439 {\color{red} (-0.00)} & 0.2960 {\color{red} (-0.01)}\\
        0-28 & 10-19 & 0.7493 {\color{blue} (+0.00)} & 0.8464 {\color{blue} (+0.00)} & 0.2969 {\color{red} (-0.00)}\\
        0-28 & 20-28 & 0.7530 {\color{blue} (+0.00)} & 0.8473 {\color{blue} (+0.00)} & 0.2984 {\color{red} (-0.00)}\\
        \midrule
        0-9 & 0-9 & 0.7730 {\color{blue} (+0.02)} & 0.8522 {\color{blue} (+0.01)} & 0.2840 {\color{red} (-0.02)}\\
        0-9 & 10-19 & 0.7620 {\color{blue} (+0.01)} & 0.8517 {\color{blue} (+0.01)} & 0.2925 {\color{red} (-0.01)}\\
        0-9 & 20-28 & 0.7788 {\color{blue} (+0.03)} & 0.8584 {\color{blue} (+0.01)} & 0.2888 {\color{red} (-0.01)}\\
        
        10-19 & 0-9 & 0.7520 {\color{blue} (+0.00)} & 0.8386 {\color{red} (-0.01)} & 0.2865 {\color{red} (-0.01)}\\
        10-19 & 10-19 & 0.7466 {\color{red} (-0.00)} & 0.8426 {\color{red} (-0.00)} & 0.2936 {\color{red} (-0.01)}\\
        10-19 & 20-28 & 0.7025 {\color{red} (-0.05)} & 0.8261 {\color{red} (-0.02)} & 0.2992 {\color{red} (-0.00)}\\
        
        20-28 & 0-9 & 0.7327 {\color{red} (-0.02)} & 0.8298 {\color{red} (-0.01)} & 0.2864 {\color{red} (-0.01)}\\
        20-28 & 10-19 & 0.7515 {\color{blue} (+0.00)} & 0.8534 {\color{blue} (+0.01)} & 0.2919 {\color{red} (-0.01)}\\
        20-28 & 20-28 & 0.6742 {\color{red} (-0.07)} & 0.8200 {\color{red} (-0.02)} & 0.3030 {\color{blue} (+0.00)}\\

        \bottomrule
        \end{tabular}
    }
    \label{tab:ab_sup_select_double_lap_layers}
\end{table}

\begin{figure*}[p]
    \centering
    \includegraphics[width =\linewidth]{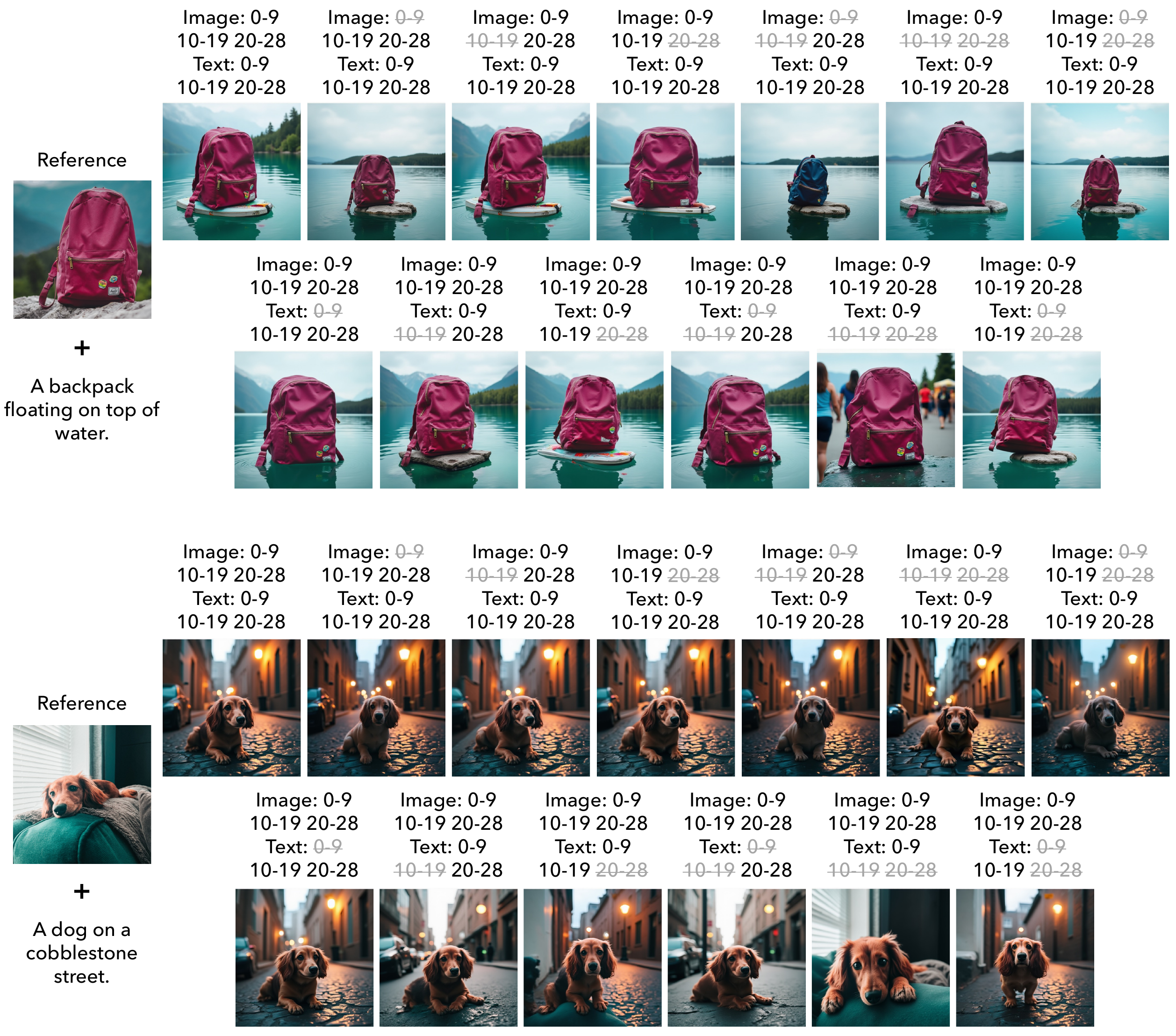}
    \caption{Qualitative results of zero-out layers for text and image modalities in our DLA module.}
\label{fig:supp_figure_inference_comparison_1}
\end{figure*}

\begin{figure*}[p]
    \centering
    \includegraphics[width =\linewidth]{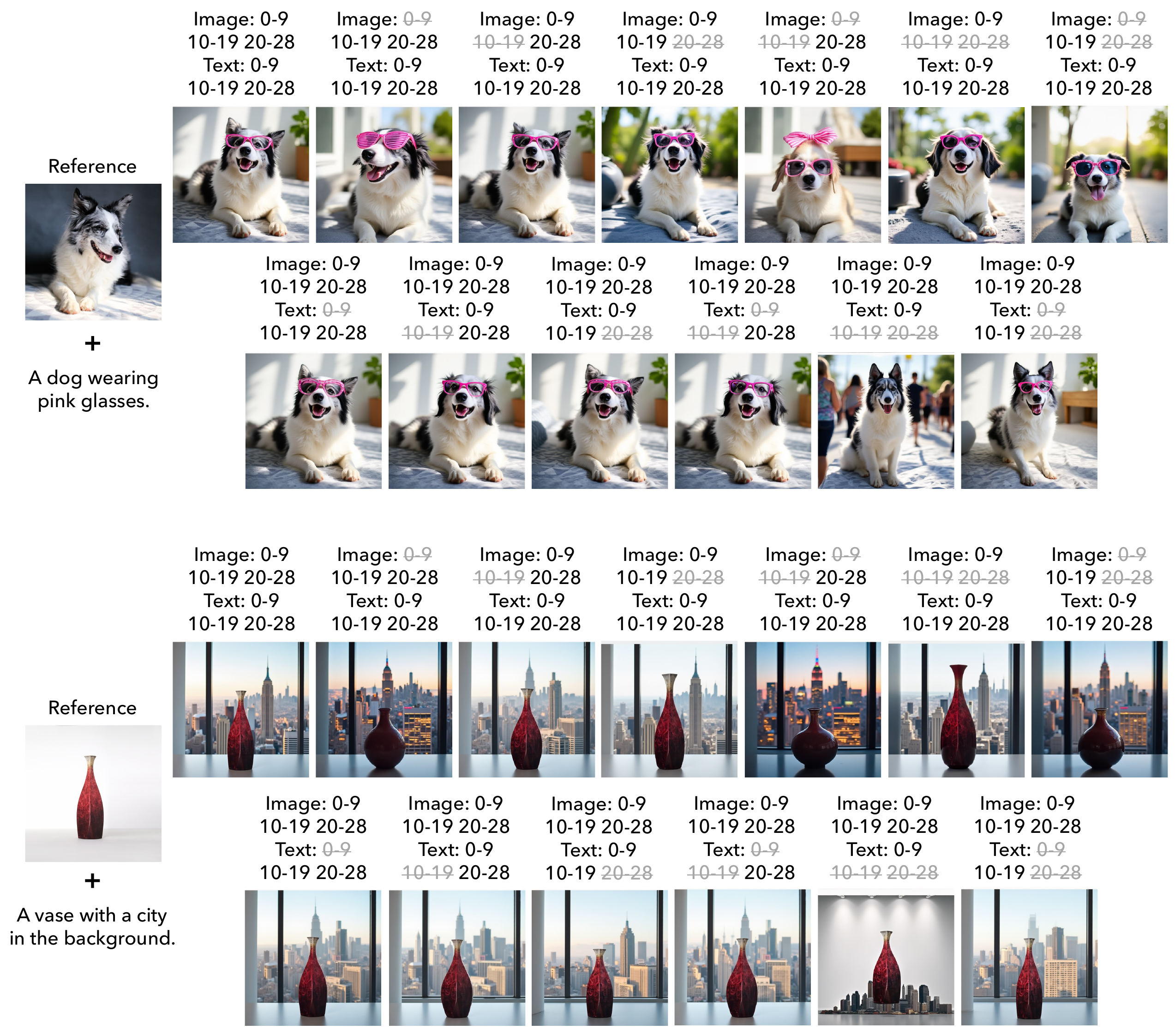}
    \caption{Qualitative results of zero-out layers for text and image modalities in our DLA module.}
\label{fig:supp_figure_inference_comparison_2}
\end{figure*}

\begin{figure*}[p]
\centering

\begin{minipage}{\linewidth}
    \centering
    \includegraphics[width=0.85\linewidth]{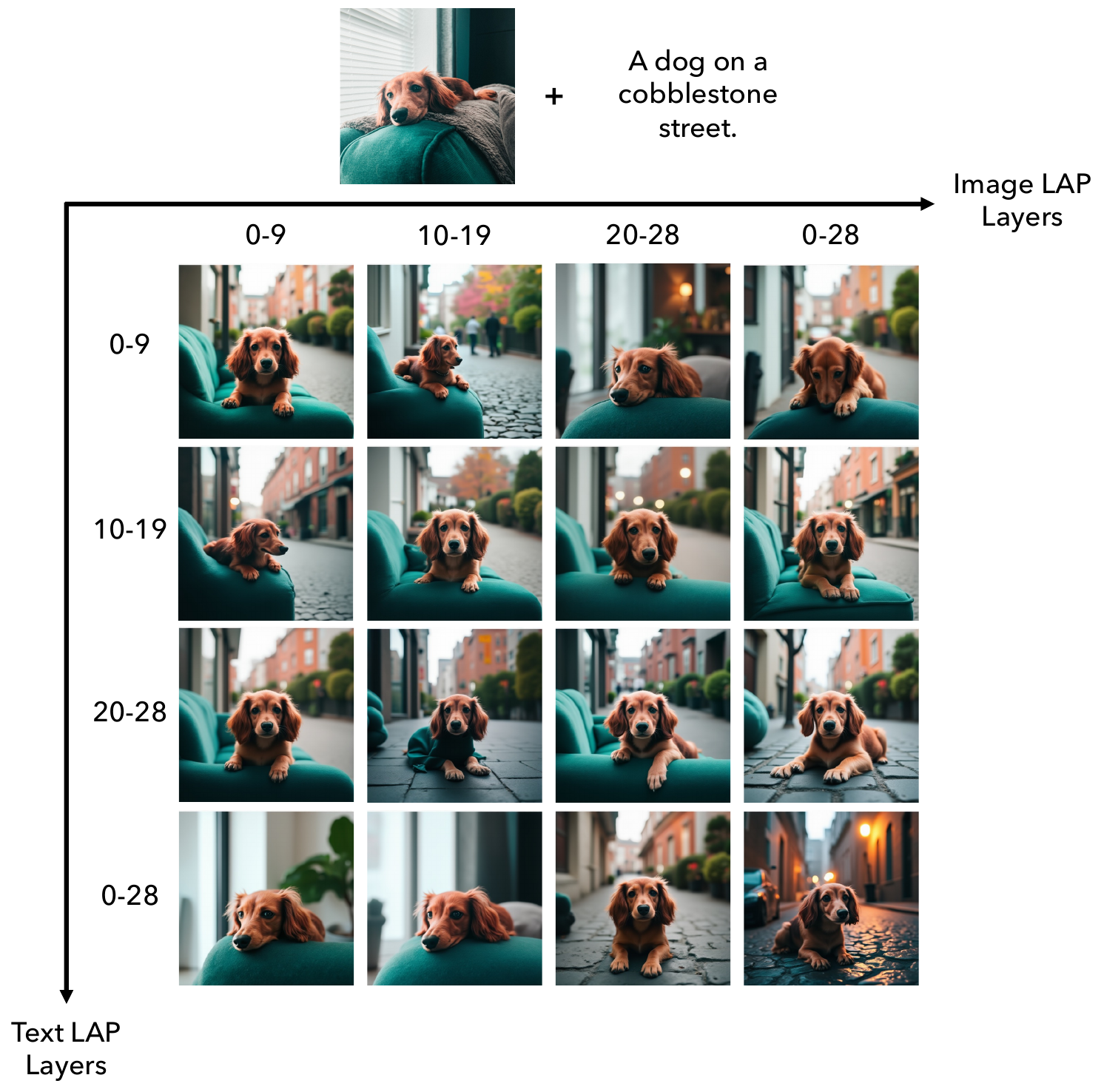}
    \captionof{figure}{Qualitative results of different layer selections for DLA. Rows correspond to text modality layer ranges, and columns correspond to image modality layer ranges, with layer groups {0–9}, {10–19}, {20–28}, and all layers (0–28). Each subplot shows the output of a separately re-trained model for the given text–image layer combination.}
    \label{fig:16figs}
\end{minipage}

\vspace{2mm}

\begin{minipage}{\linewidth}
    \centering
    \includegraphics[width=0.8\linewidth]{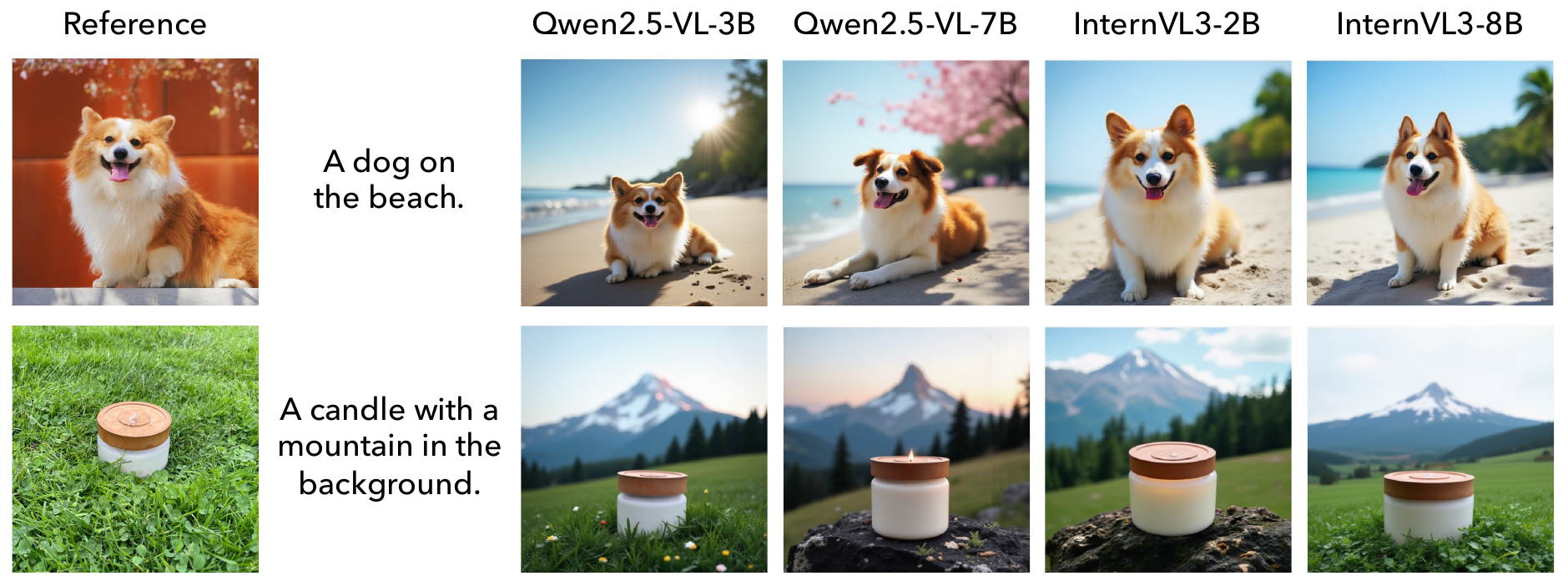}
    \captionof{figure}{Qualitative comparison of different MLLM backbones, including Qwen2.5-VL-3B~\cite{bai2025qwen25vltechnicalreport}, Qwen2.5-VL-7B~\cite{bai2025qwen25vltechnicalreport}, InternVL3-2B~\cite{zhu2025internvl3}, and InternVL3-8B~\cite{zhu2025internvl3}.}
    \label{fig:qwen_vs_internvl}
\end{minipage}

\end{figure*}

We present a qualitative comparison of different layer selection strategies for the DLA module in Figure~\ref{fig:16figs}. The figure shows 16 combinations of layers for the text and image modalities, including ranges 0-9, 10-19, 20-28, and all layers (0-28). Each row corresponds to the text modality layer setting, while each column represents the image modality layer setting. Importantly, each combination represents a separately re-trained model, allowing us to isolate the effect of specific layer selections on the final generation. Compared with the inference-time zero-out analysis in Figure~\ref{fig:supp_figure_inference_comparison_1} and Figure~\ref{fig:supp_figure_inference_comparison_2}, we observe that pre-selecting layers during training can lead to serious relaying on identity image and worse text-following capability as shown in Figure~\ref{fig:16figs}, especially when not using all 28 layers for the text modality. This visualization highlights the trade-offs between constraining layers for efficiency and maintaining balanced identity preservation and multimodal understanding.

\section{Additional sensitivity analysis}
\label{sec:sensitivity}

\begin{figure}[t]
\centering

\begin{minipage}[t]{0.52\textwidth}
\vspace{0pt}
    \centering
    \includegraphics[width =0.8\linewidth]{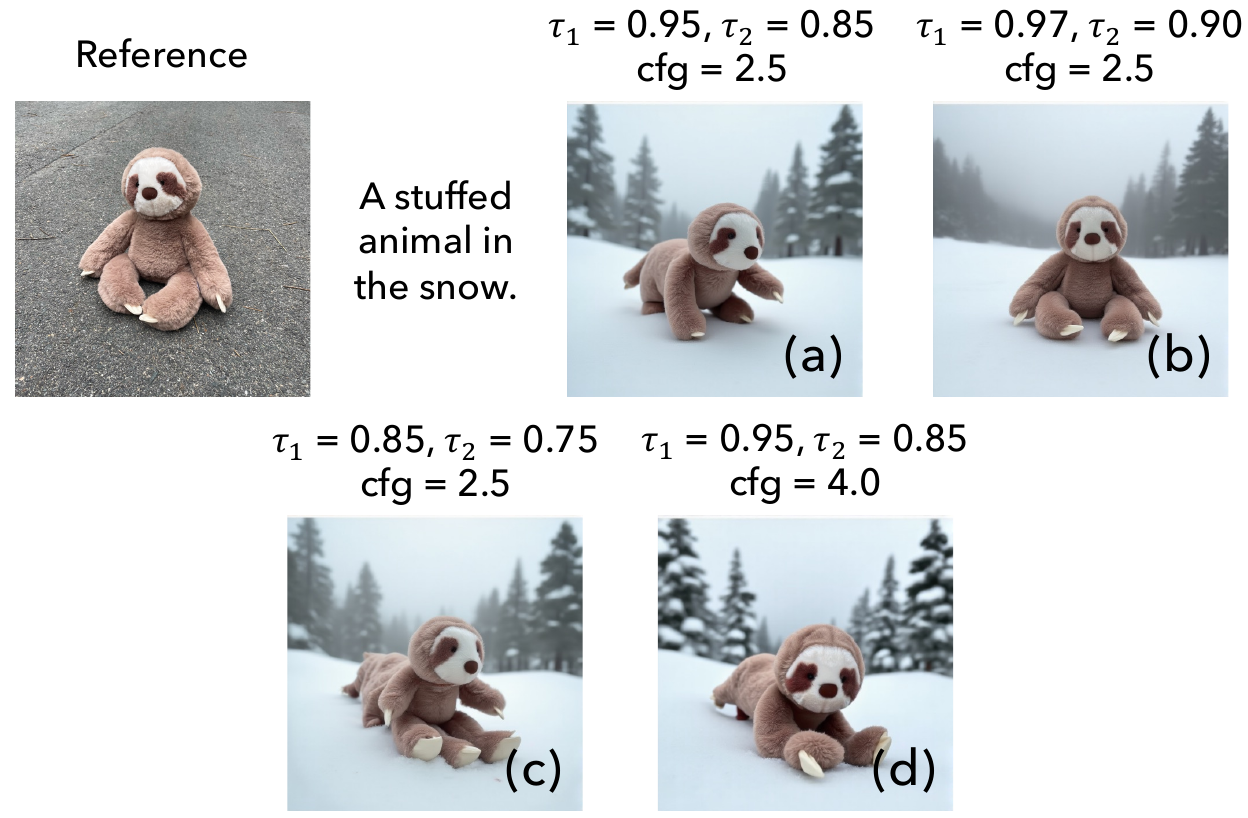}
    \caption{Ablations on different configurations of the timestep-aware denoising process. The thresholds $\tau_1$ and $\tau_2$, which partition the denoising stages, along with the CFG value, can be adjusted freely by users to balance high-fidelity identity preservation with diverse pose generation.}
    \label{fig:ms_denoising_qual}
    \vspace{-5pt}
\end{minipage}
\hfill
\begin{minipage}[t]{0.46\textwidth}
\vspace{5pt}
\captionof{table}{Quantitative results for different multi-stage denoising configurations, with the gray row indicating our current parameter choice. The thresholds $\tau_1$ and $\tau_2$ along with the CFG value control the trade-off between identity preservation and text following, while the model remains robust across a range of parameter settings.}
\rowcolors{2}{white}{uoftcoolgray!25}
    \centering
    \resizebox{0.9\linewidth}{!}{
        \begin{tabular}{cccccc}
        \toprule
           \textbf{$\tau_1$}  & \textbf{$\tau_2$} & \textbf{CFG} &  \textbf{DINO-I ($\uparrow$)}  & \textbf{CLIP-I ($\uparrow$)} & \textbf{CLIP-T ($\uparrow$)}  \\ 
        \midrule
        0.00 & 0.00 & 2.5 & 0.6905 & 0.8225 & \textbf{0.3044}\\ 
        1.00 & 1.00 &  2.5 & 0.7351 & 0.8462 & 0.2554\\ 
        0.97 & 0.90 & 2.5 & \textbf{0.7490} & \textbf{0.8466} & 0.2963\\ 
        0.85 & 0.70 & 2.5 & 0.7282 & 0.8376  & 0.3034\\ 
        0.95 & 0.85 & 1.5 & 0.7268 & 0.8385 & 0.2990\\ 
        0.95 & 0.85 & 2.0 & 0.7418 & 0.8430 & 0.3004 \\ 
        \cellcolor{gray!50}0.95 & \cellcolor{gray!50}0.85 & \cellcolor{gray!50}2.5 & \cellcolor{gray!50}0.7482 &  \cellcolor{gray!50}0.8443&  \cellcolor{gray!50}0.3010\\ 
        0.95 & 0.85 & 3.0   & 0.7481 & 0.8428 & 0.3008\\ 
        0.95 & 0.85 & 4.0   & 0.7431 & 0.8381 & 0.3012\\ 
        \bottomrule
        \end{tabular}
    }
    \label{tab:timestep_aware}

    
\end{minipage}
\end{figure}

Our framework, which leverages both MLLM and VAE for identity preservation, supports a multi-stage, timestep-aware denoising process: early steps rely on MLLM features for high-level reasoning, while later steps use VAE features for fine-grained detail. We ablate different configurations of this process to guide users in selecting denoising thresholds ($\tau_1, \tau_2$) that balance identity fidelity and pose variation. As shown in Figure~\ref{fig:ms_denoising_qual}, higher thresholds (b) improve identity preservation but reduce pose diversity, whereas lower thresholds (c) allow more creative poses, but with slightly compromised identity. Sample (d) illustrates that extreme CFG values can degrade image quality. This design offers users flexibility to control the trade-off between subject fidelity and creativity, and Table~\ref{tab:timestep_aware} shows that the overall performance remains robust across a range of parameter choices.

\section{Ablation on different MLLM selection}
\label{sec:mlmm_ablation}
For our main evaluation, we use InternVL3-8B~\cite{zhu2025internvl3} as it provides a good balance between model capacity and efficiency. To study the impact of different MLLM backbones, we further ablate a range of alternatives with varying sizes and architectures, including InternVL3-2B~\cite{zhu2025internvl3}, Qwen2.5-VL-3B~\cite{bai2025qwen25vltechnicalreport}, and Qwen2.5-VL-7B~\cite{bai2025qwen25vltechnicalreport}. As shown in Table~\ref{tab:different_mllm_selection}, all models follow a similar performance trend across the metrics, with only minor variations. Qwen2.5-VL exhibits slightly weaker visual context understanding but marginally better text alignment. Overall, the difference is not substantial. Notably, InternVL3-2B achieves comparable results to the 8B model while using significantly fewer parameters, offering a promising lightweight alternative.

We present a qualitative comparison of different MLLM backbones in Figure~\ref{fig:qwen_vs_internvl}. Although these models differ in type and parameter size, the results do not reveal any major visual differences across the backbones. All four models, including Qwen2.5-VL-3B, Qwen2.5-VL-7B, InternVL3-2B, and InternVL3-8B, produce results with comparable image fidelity, identity preservation, and text-following ability. While larger models show slightly stronger grounding and semantic alignment, the overall performance trend remains consistent, indicating that our DLA framework generalizes well across different MLLM architectures. This further supports the conclusion from the quantitative analysis that the choice of backbone has limited impact on the final generation quality.

\begin{table}[t]
    \rowcolors{2}{white}{uoftcoolgray!25}
    \centering
    \caption{Ablation on different MLLM backbones. We compare InternVL3-8B—used as our main model—with alternative architectures of varying sizes, including InternVL3-2B, Qwen2.5-VL-3B, and Qwen2.5-VL-7B.}
    \vspace{5pt}
    \resizebox{0.6\linewidth}{!}{
        \begin{tabular}{lccc}
        \toprule
        \textbf{Model} &  \textbf{DINO-I ($\uparrow$)} & 
        \textbf{CLIP-I ($\uparrow$)} & 
        \textbf{CLIP-T ($\uparrow$)} \\ 
        \midrule
        InternVL3-8B & 0.7482 & 0.8443 & 0.3010\\
        \midrule
        InternVL3-2B & 0.7415 & 0.8380 & 0.2987\\
        Qwen2.5-VL-3B & 0.7194 & 0.8300 & 0.3027\\ 
        Qwen2.5-VL-7B & 0.7282 & 0.8241 & 0.3031\\
        \bottomrule
        \end{tabular}
    }
    \label{tab:different_mllm_selection}
\end{table}

\section{Adaptation to multi-subject generation}
\label{sec:multi_subject}
\begin{figure*}[p]
    \centering
    \includegraphics[width =\linewidth]{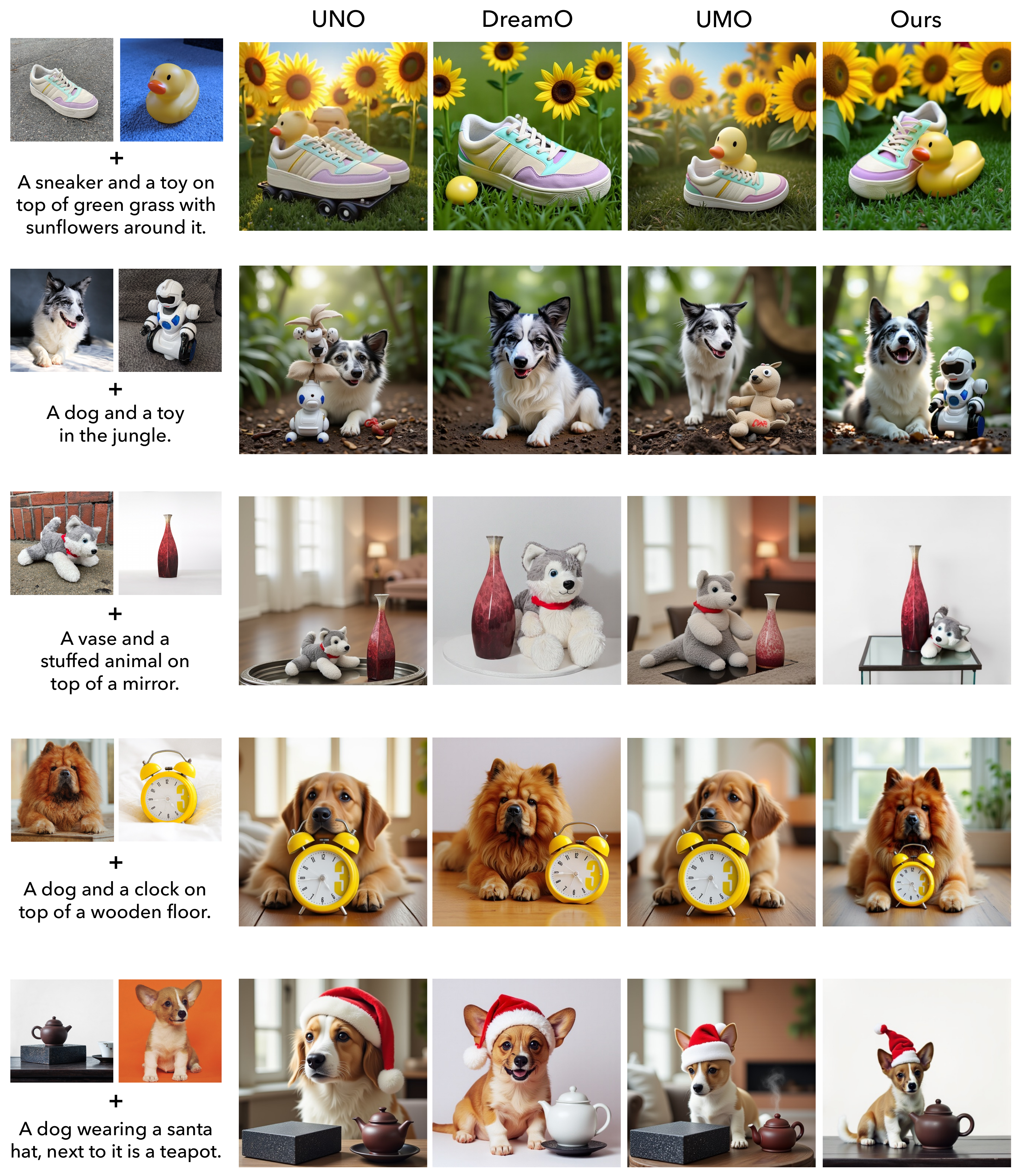}
    \caption{Multi-reference generation results. Although our method is originally designed for single-subject generation, it adapts effectively to multi-subject scenarios after lightweight fine-tuning on MUSAR-Gen~\cite{guo2025musar}. Compared to UNO~\cite{wu2025less}, DreamO~\cite{mou2025dreamo}, and UMO~\cite{cheng2025umo}, our model achieves clearer identity separation, consistent posture, and more reliable concept binding across subjects.}
\label{fig:supp_figure_multi_subject}
\end{figure*}

In the main paper, we primarily focus on single-subject generation. We intentionally focus on single-subject training for two reasons: (1) our primary goal is to explore and analyze how to optimally leverage MLLM features for subject-driven generation; and (2) high-quality multi-subject datasets are often private and difficult to obtain at scale. However, our framework can also be extended to handle multi-subject generation with minimal adaptation. Hence, we fine-tune the model using the public two-subject dataset MUSAR-Gen~\cite{guo2025musar}, which contains fewer than 30K image pairs. During training, after completing the MLLM-only stage on UNO-1M~\cite{wu2025less} for single-subject learning, we continue to fine-tune the model on MUSAR-Gen—still within the MLLM-only framework. In the subsequent stage involving both MLLM and VAE, we jointly train on a mixture of UNO-1M and MUSAR-Gen to establish the full multi-subject pipeline. We compare the resulting multi-subject model against UNO~\cite{wu2025less}, DreamO~\cite{mou2025dreamo}, and UMO~\cite{cheng2025umo}. As shown in Figure~\ref{fig:supp_figure_multi_subject}, our method achieves superior results on the multi-subject DreamBench~\cite{ruiz2023dreambooth} samples, excelling in both identity preservation and text compliance.

\section{Multimodal reasoning benchmark construction}
\label{sec:mm_reasoning_construction}

In the main paper, to quantify the multimodal reasoning capability, we propose to evaluate on a constructed benchmark consisting of complex prompts that require the model to perform concept binding and cross-modal reasoning. The key idea for constructing this benchmark is to collect images where a primary subject appears together with additional visible objects that function as accessories. The corresponding text prompts, however, deliberately refer to a non-salient object in the image rather than the main subject. Under this setting, the models cannot simply presume that the most salient object in the reference image is the subject whose identity needs to be preserved. Instead, they are expected to reason about the prompt and correctly locate the concept mentioned in the text within the image.

To curate such images, we collect generated samples from state-of-the-art subject-driven methods like USO on DreamBench, and manually verify their content and quality. The associated prompts are then modified by replacing the subject category in the original prompts. For example, the prompt ``A \textit{cat} wearing a shirt'' can be changed to ``An \textit{elephant} wearing a shirt''. Furthermore, we construct variants that include two reference images. Each sample is formed by combining a generated composite image, an original reference image from DreamBench, and a modified prompt in which the subject category matches that of the selected DreamBench image. Samples in the curated benchmark can be seen in Figure~\ref{fig:supp_figure_mm_samples}. In total, the benchmark contains 170 single-reference samples and 180 two-reference samples, resulting in 350 test samples overall.

\section{Additional implementation details for reproducibility}
\label{sec:additional_implementation}

\subsection{Model architecture and trainable parameters}

\textbf{Training setup.}
We initialize the diffusion model from
\texttt{black-forest-labs/FLUX.1-dev} and load the multimodal large language model together with its processor from \texttt{OpenGVLab/InternVL3-8B-hf}. The original T5
text conditioning feature for the diffusion model is replaced by the feature from the MLLM. We retain
the pooled text embedding from the frozen
\texttt{openai/clip-vit-large-patch14} encoder, whose tokenizer uses a
maximum length of 77 with the pooled vector having 768 channels. The MLLM, VAE, and CLIP encoder are frozen.

\textbf{Dual layer aggregation.}
We collect the input embedding and all transformer hidden states as the input to the dual layer aggregator (DLA) module. The selected InternVL3-8B model with $M=28$ transformer layers thus yields $K=M+1=29$ input features to the DLA module. Text and image tokens are processed by two
independently parameterized LAP modules. Each module contains one
pre-norm self-attention block with 32 heads, LayerNorm with
$\epsilon=1\text{e-}6$, and a GELU feed-forward network. Residual connections inside this block are retained. After processing by the self-attention blocks, the output is reduced over the number of features by a learned linear mapping $\mathbb R^{K}\to\mathbb R$ to integrate the $K$ MLLM features into one for conditioning the diffusion model with an alignment projector initialized with Xavier-uniform linear weights, zero bias, and unit LayerNorm scales.

\textbf{LoRA placement.}
We add rank-512 LoRA modules to the image/text QKV, the output
projection layer in all the single-stream and double-stream blocks in Flux for efficient finetuning. The LoRA modules use
unit update scaling with no alpha multiplier and no adapter dropout. The down-projection weights initialized from
$\mathcal N(0,(1/512)^2)$, while the up-projection weights are initialized with zero.

\subsection{Input data preprocessing}

Each single-subject example contains the reference
image along with the prompt text. We use the native chat
template with padding, tokenization, and an appended
prompt stating that our current task is image generation. No generated answer tokens are required. The model returns all hidden states in one forward pass. Image-token positions
are identified from token names supplemented by explicit image token IDs.
The rest of the tokens, including template and padding positions, are
processed by the text LAP.

The reference images in MLLM retain square padding and resizing, while the references and target images for the VAE input instead use aspect-ratio bucketing for preprocessing. Specifically, the $(H,W)$ buckets are
\[
\begin{gathered}
(256,768),(320,768),(320,704),(384,640),(448,576),(512,512),\\
(576,448),(640,384),(704,320),(768,320),(768,256).
\end{gathered}
\]
The reference images will be resized to the selected bucket that minimizes the
absolute difference from each batch's $H/W$ ratio.

\subsection{Dropout strategy}

As mentioned in Section~\ref{sec:denoising}, during training, dropout will be implemented on the input modalities to ensure robostness and the capability of performing multi-stage timestep-aware denoising where only part of the conditioning signals may exist in each denoising stage. In implementation, each sample independently decide on whether the text and image for the MLLM and the VAE reference to be dropped, with a probability of 0.1. Dropped MLLM text is replaced
by an empty string, while dropped MLLM images are replaced by a black dummy image to preserve the structure of the message.
Dropped VAE conditioning will be zeroed after encoding, retaining
its token positions.

\subsection{DreamBench metric computation}

Following the official implementation in DreamBooth~\cite{ruiz2023dreambooth}, we use CLIP ViT-B/32 and DINO ViT-S/16 for calculating the scores of DINO-I, CLIP-I, and CLIP-T. Both reference and generated
images are first resized to $224\times224$ using bicubic
interpolation, and then passed through the respective model
preprocessors. DINO-I uses the L2-normalized final \texttt{CLS} token to obtain the cosine similarity between the features from the reference image and the generated image, while
CLIP-I and CLIP-T use normalized image and text embeddings for similarity calculation.

\section{Behind the scenes}
\label{sec:unsuccessful_attempts}

We would like to share about the "behind the scenes" information about how our project developed and how some unsuccessful explorations informed our final design over time. Originally, we aim at investigating whether MLLM features alone could provide sufficient reference image conditioning for subject-driven generation. This would simplify the framework by allowing a unified multimodal representation to support both multimodal understanding and identity preservation. We explored various types of MLLM-only conditioning strategies, including the variants in Tables~\ref{tab:ab1_strategy} and~\ref{tab:ab_sup_select_double_lap_layers}, but found out that solely using MLLM features for diffusion model conditioning is not enough to achieve as good subject identity preservation performance as using VAE features. 

These findings motivated us to propose our final model as an MLLM-VAE framework with the multi-stage denoising process, which combines the complementary strengths of both the MLLM and the VAE. MLLM features guide generation with their strong multimodal understanding capability during the early and middle denoising stages, while VAE features help recover fine-grained appearance details to preserve subject identity. This design does not undermine the crucial role of the MLLM in our subject-driven generation framework. Instead, it reveals the benefits and limitations of both MLLMs and VAEs in subject-driven generation, and further provides guidance on how these two models can be collaboratively applied towards the goal of both multimodal understanding and identity preservation in the subject-driven generation task.

\section{User study}
\label{sec:user_study}

Below, we provide further details of the user study setup. Participants are given the following instructions at the beginning of the study.

\vspace{15pt}

\hrule
\vspace{5pt}
\textit{In subject-driven generation, the user gives a reference image along with a text prompt, and the goal is to generate an image that \textbf{aligns with the text description while preserving the identity in the reference image}.}

\textit{For each of the following cases, we use 5 different methods to perform subject-driven generation. We would like to invite you to give an overall score from 1 (worst quality) to 10 (best quality) to measure the quality of the generated results.}

\textit{There are a few points to consider when providing the scores:}
\begin{itemize}
    \item \textit{Whether the generated images preserve the identity of the reference image}
    \item \textit{Whether the generated images follow the text prompt}
    \item \textit{The visual quality of the generated images (e.g., whether they look realistic, whether they follow the physical rules)}
\end{itemize}
\vspace{5pt}
\hrule

\vspace{15pt}

\begin{figure*}[p]
    \centering
    \includegraphics[width =\linewidth]{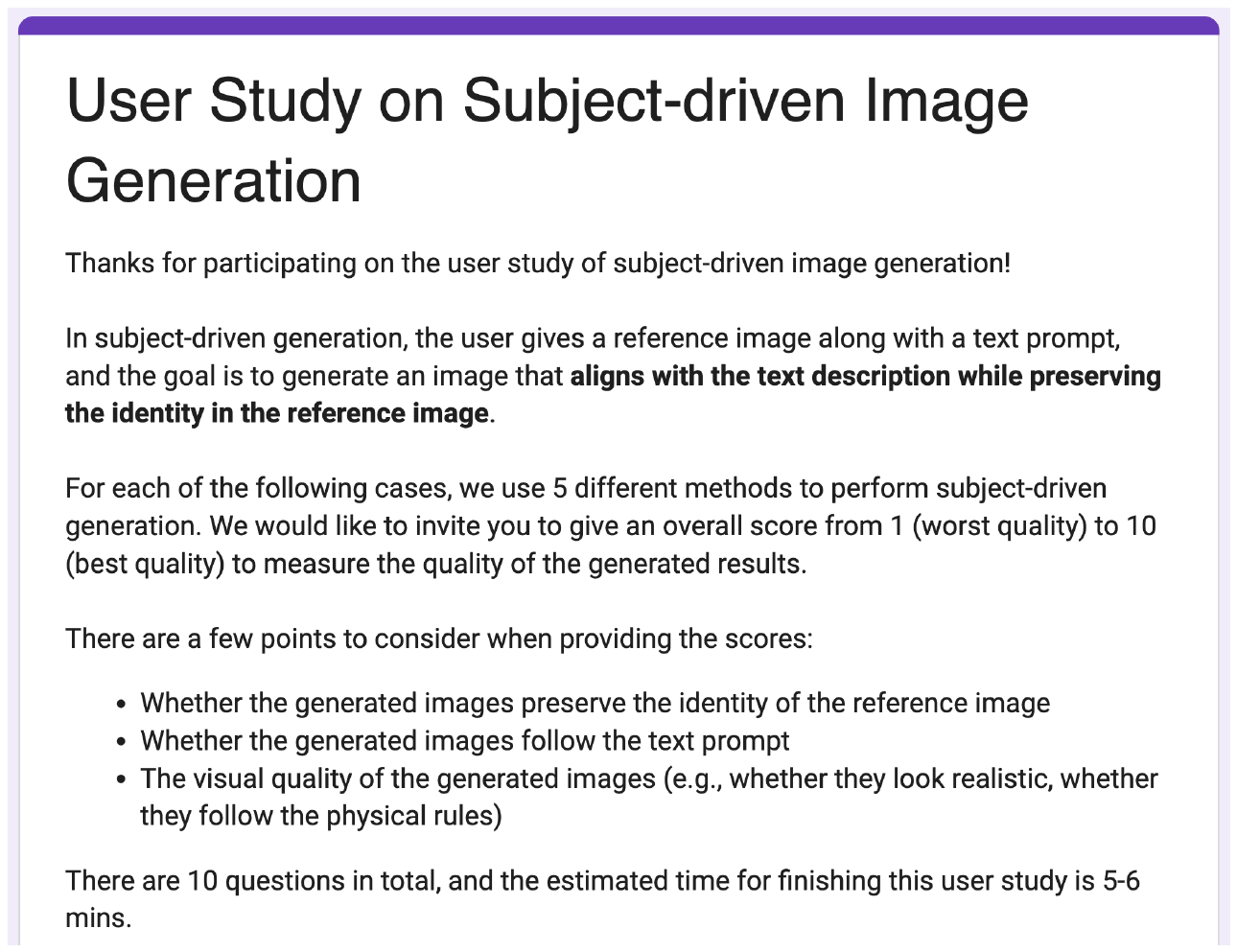}
    \caption{Screenshot of the instructions of our user study.}
\label{fig:supp_figure_user_study_1}
\end{figure*}

\begin{figure*}[p]
    \centering
    \includegraphics[width =0.9\linewidth]{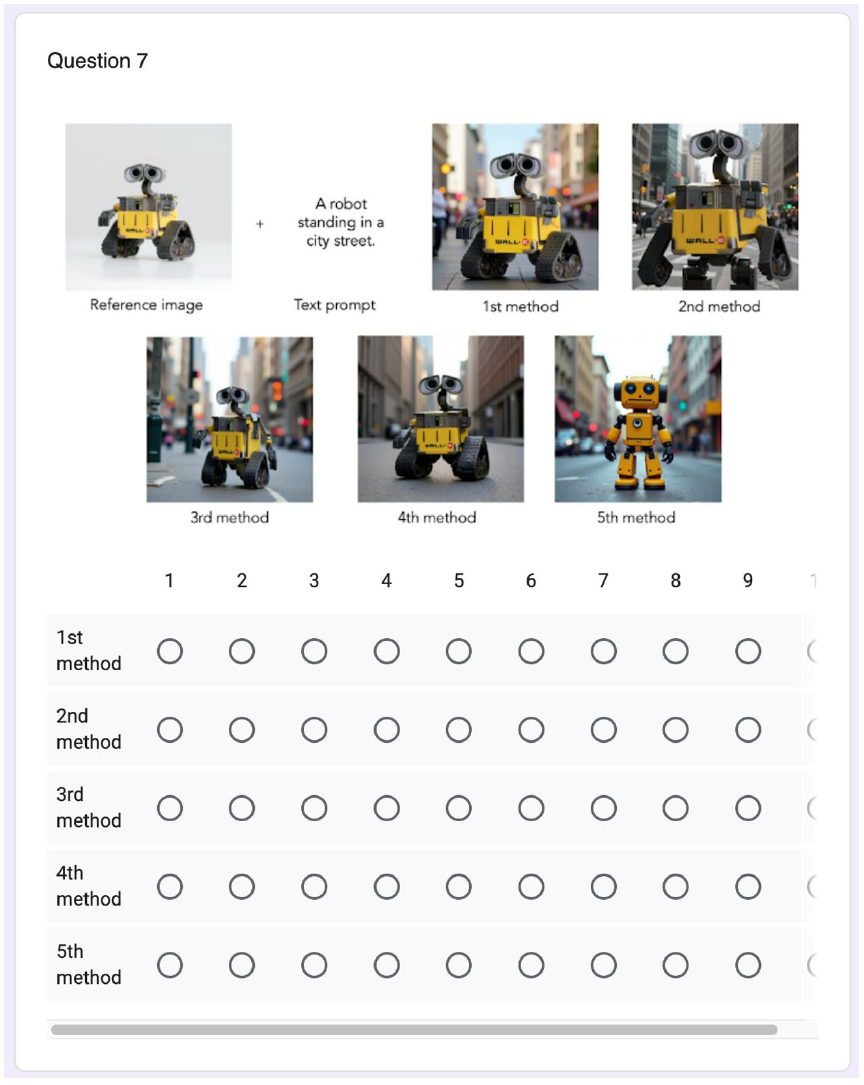}
    \caption{Screenshot of the user study interface with an example question, containing the reference image, text prompts, and images generated by five methods placed side by side \textit{in random order}, and the questions to score the five generated results.}
\label{fig:supp_figure_user_study_2}
\end{figure*}

We also include screenshots of the user study interface in Figures~\ref{fig:supp_figure_user_study_1} and~\ref{fig:supp_figure_user_study_2}.

\section{MLLM-based evaluation details}
\label{sec:vlm_based_evaluation}

We follow the concept preservation evaluation protocol in DreamBench++ \cite{peng2025dreambench} and its follow-ups~\cite{zheng2025track,kumar2025deft} to construct prompts for MLLM-based scoring of identity preservation between the reference image and the generated image. The prompts are provided below.


\vspace{15pt}

\hrule

\vspace{5pt}

\noindent\textbf{\textit{\#\#\# Task Definition}} \\[0.1pt]

\noindent\textit{You will be provided with an image generated based on reference image.} \\[0.1pt]

\noindent\textit{As an experienced evaluator, your task is to evaluate the semantic consistency between the subject of the generated image and the reference image, according to the scoring criteria.} \\[0.1pt]

\noindent\textbf{\textit{\#\#\# Scoring Criteria}} \\[0.1pt]

\noindent\textit{It is often compared whether two subjects are consistent based on four basic visual features:}

\begin{enumerate}
    \item \textit{Shape: Evaluate whether the main body outline, structure, and proportions of the generated image match those of the reference image. This includes the geometric shape of the main body, clarity of edges, relative sizes, and spatial relationships between various parts composing the main body.}
    \item \textit{Color: Comparing the accuracy and consistency of the main colors generated in the image with those of the reference image. This includes saturation, hue, brightness, and whether the distribution of colors is similar to that of the subject in the reference image.}
    \item \textit{Texture: Focus on the local parts of the RGB image, whether the generated image effectively captures fine details without appearing blurry, and whether it possesses the required realism, clarity, and aesthetic appeal. Please note that unless specifically mentioned in the text prompt, excessive abstraction and formalization of texture are not necessary.}
    \item \textit{Facial Features: If the evaluation is of a person or animal, facial features will greatly affect the judgment of image consistency, and you also need to focus on judging whether the facial area looks very similar visually.}
\end{enumerate}

\noindent\textbf{\textit{\#\#\# Scoring Range}} \\[0.1pt]

\noindent\textit{You need to give a specific integer score based on the comprehensive performance of the visual features above, ranging from 0 to 4:}

\begin{itemize}
    \item \textit{Very Poor (0): No resemblance. The generated image's subject has no relation to the reference.}
    \item \textit{Poor (1): Minimal resemblance. The subject falls within the same broad category but differs significantly.}
    \item \textit{Fair (2): Moderate resemblance. The subject shows likeness to the reference with notable variances.}
    \item \textit{Good (3): Strong resemblance. The subject closely matches the reference with only minor discrepancies.}
    \item \textit{Excellent (4): Near-identical. The subject of the generated image is virtually indistinguishable from the reference.}
\end{itemize}

\noindent\textbf{\textit{\#\#\# Input Format}} \\[0.1pt]

\noindent\textit{Every time you will receive two images, the first image is a reference image, and the second image is the generated image.} \\[0.1pt]

\noindent\textit{Please carefully review each image of the subject.} \\[0.1pt]

\noindent\textbf{\textit{\#\#\# Output Format}} \\[0.1pt]

\noindent\textit{Score: [Your Score]} \\[0.1pt]

\noindent\textit{You must adhere to the specified output format, which means that only the scores need to be output, excluding your analysis process.}

\vspace{5pt}
\hrule





\begin{table}[h]
    \rowcolors{2}{white}{uoftcoolgray!25}
    \caption{Quantitative results on additional benchmarks of XVerseBench~\cite{chen2025xverse} and LAMICBench~\cite{chen2025lamic}, both on the IP-Sim metric, where higher value means better performance.}
    \centering
    \vspace{5pt}
    \resizebox{0.7\linewidth}{!}{
        \begin{tabular}{lcccc}
        \toprule
        Dataset  & DreamO & UNO &
        USO  & Ours \\ 
        \midrule
        XVerseBench  & 76.08 & \textbf{80.36} & 78.90   & \underline{79.10}\\
        LAMICBench (two-subject)  & \underline{65.25} & 64.93 & Not Applicable   & \textbf{66.46}\\
        \bottomrule
        \end{tabular}
    }
\label{tab:numbers_in_additional_benchmarks}
\end{table}

\section{Evaluation on more benchmarks}
\label{sec:more_benchmarks}

We include evaluation on additional benchmarks of XVerseBench~\cite{chen2025xverse} and LAMICBench~\cite{chen2025lamic}. Specifically, we conduct experiments on the single-subject set of XVerseBench and the two-reference subset of LAMICBench with a slightly finetuned version of our model trained on two-subject data as described in Section~\ref{sec:multi_subject}, both on the subject categories. The results, shown in Table~\ref{tab:numbers_in_additional_benchmarks}, demonstrate the decent performance of our method. Please note that we refrained from including human face and identity in our evaluation because our model is not trained on human-related data due to ethical considerations.

\section{More qualitative results}
\label{sec:qualitative}
\subsection{Stress testing}

\begin{figure*}[p]
\centering

\begin{minipage}{\linewidth}
    \centering
    \includegraphics[width=0.8\linewidth]{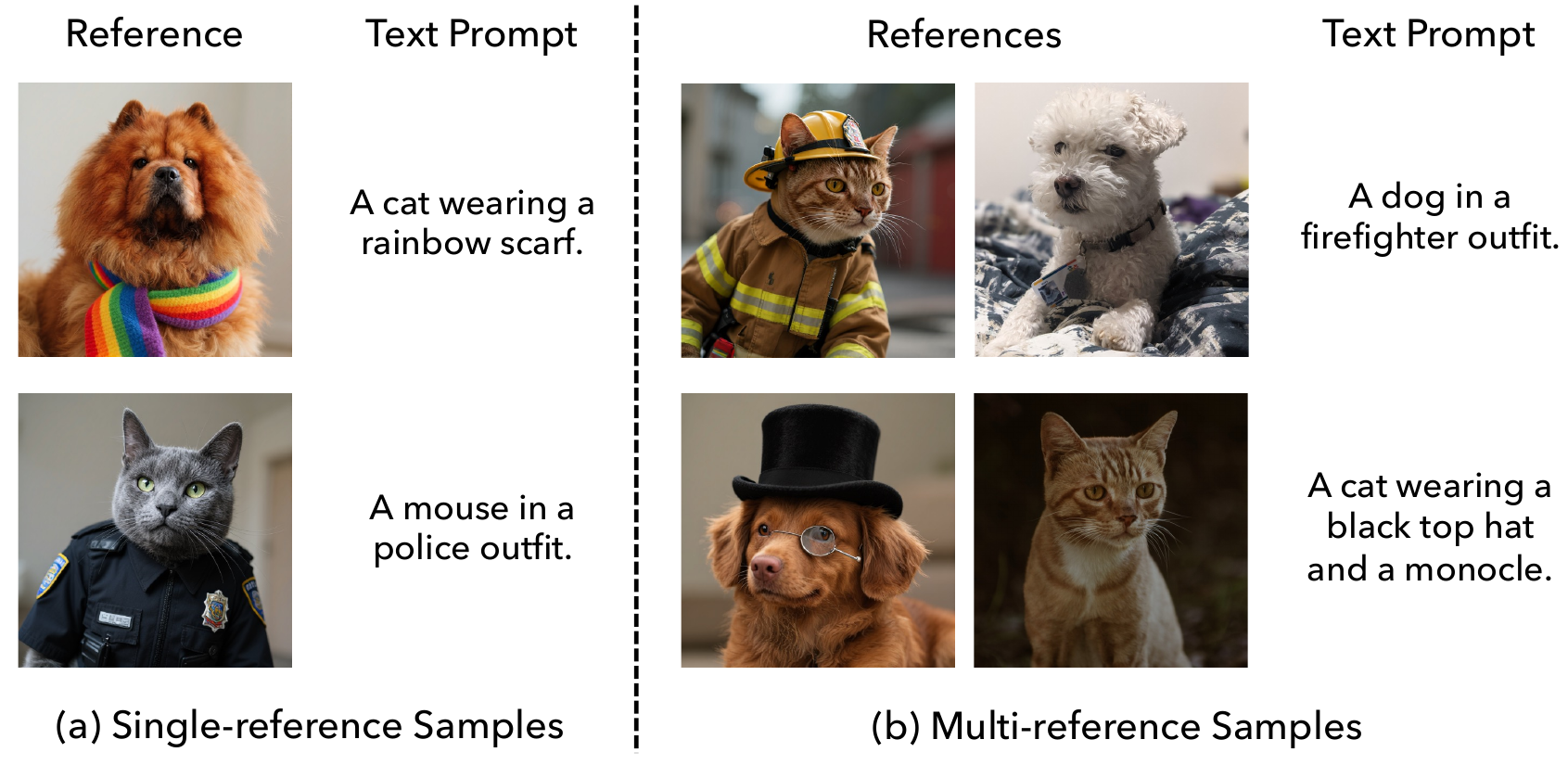}
    \captionof{figure}{Test samples from the constructed multimodal reasoning benchmark.}
    \label{fig:supp_figure_mm_samples}
\end{minipage}

\vspace{6mm}

\begin{minipage}{\linewidth}
    \centering
    \includegraphics[width=0.85\linewidth]{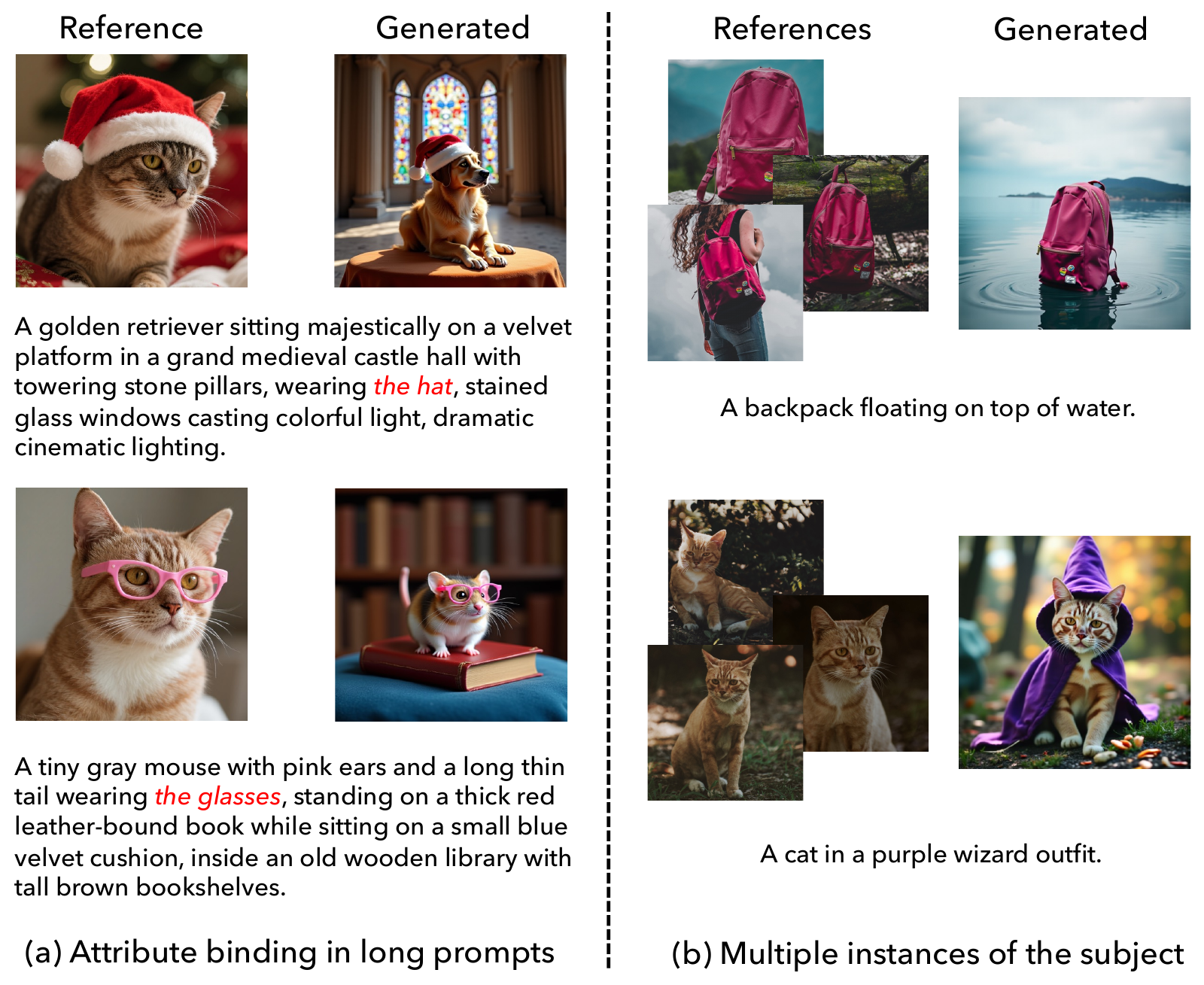}
    \captionof{figure}{Stress testing performance of our method on challenging scenarios, including attribute binding in long context prompts, and test cases that contain multiple instances of the subject. The results demonstrate the robustness of the strong reasoning capability within our MLLM-DiT system.}
    \label{fig:supp_figure_stress_testing}
\end{minipage}

\end{figure*}

To further evaluate the robustness of our method under more challenging conditions, we provide additional stress test samples involving complex instructions. In particular, we consider two scenarios: (1) prompts containing multiple instances of the subject, and (2) attribute binding in long context prompts. These settings require the model to correctly preserve subject identity while simultaneously satisfying multiple constraints specified in the text. As shown in Figure~\ref{fig:supp_figure_stress_testing}, our method has the ability of handling multiple instances and bind the concepts in complex prompts, showing the robustness and the strong reasoning capability within our MLLM-DiT system.

\subsection{Additional qualitative comparisons}
\begin{figure*}[p]
    \centering
    \includegraphics[width =\linewidth]{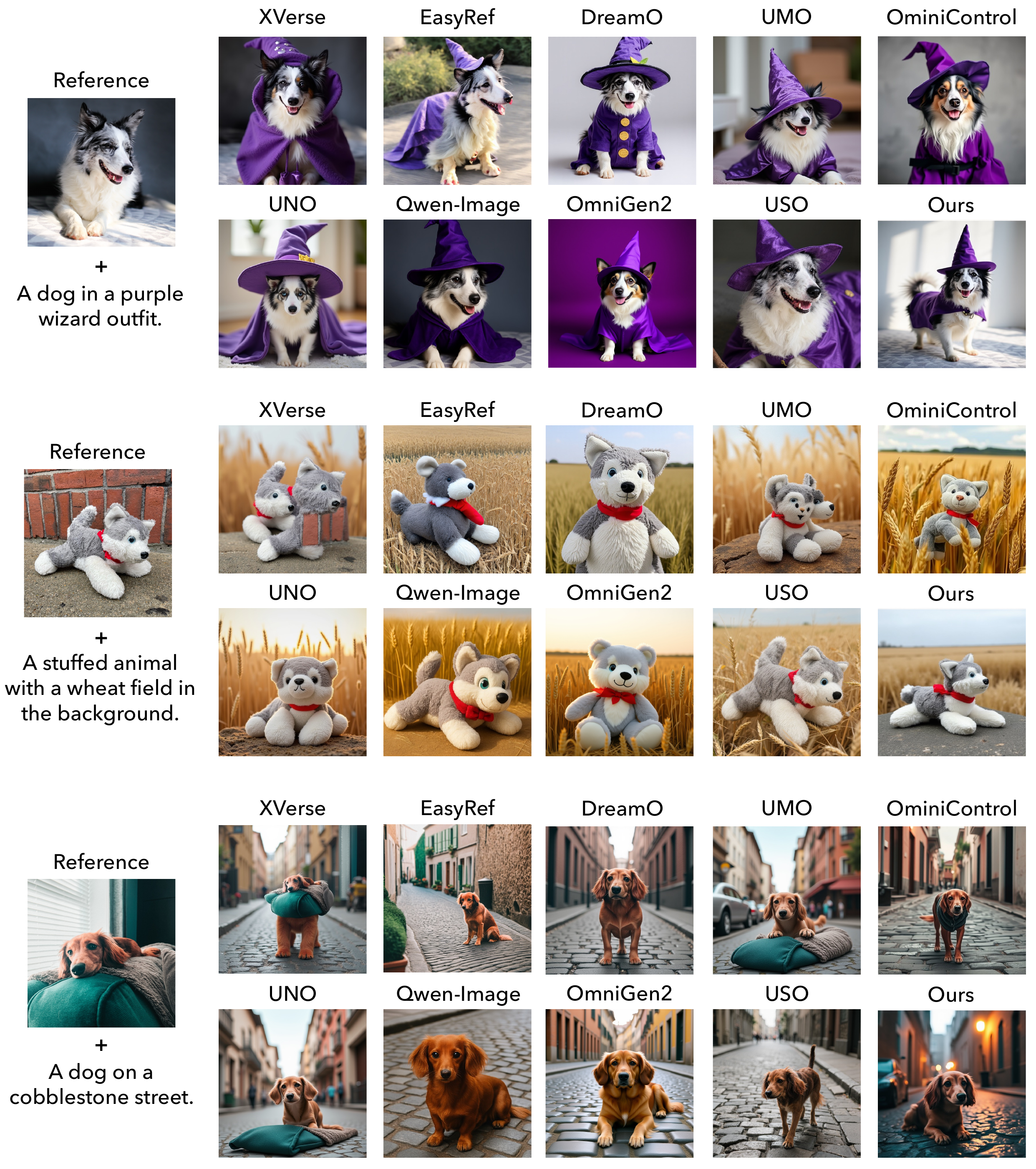}
    \caption{Additional qualitative comparisons with state-of-the-art subject-driven generation methods. Our method consistently achieves better identity preservation and text alignment across diverse prompts. We compare against XVerse~\cite{chen2025xverse}, EasyRef~\cite{zong2024easyref}, DreamO~\cite{mou2025dreamo}, UMO~\cite{cheng2025umo}, OminiControl~\cite{tan2024ominicontrol}, UNO~\cite{wu2025less}, Qwen-Image~\cite{wu2025qwen}, OmniGen2~\cite{wu2025omnigen2}, and USO~\cite{wu2025uso}.}
\label{fig:supp_figure_more_comparison_1}
\end{figure*}

\begin{figure*}[p]
    \centering
    \includegraphics[width =\linewidth]{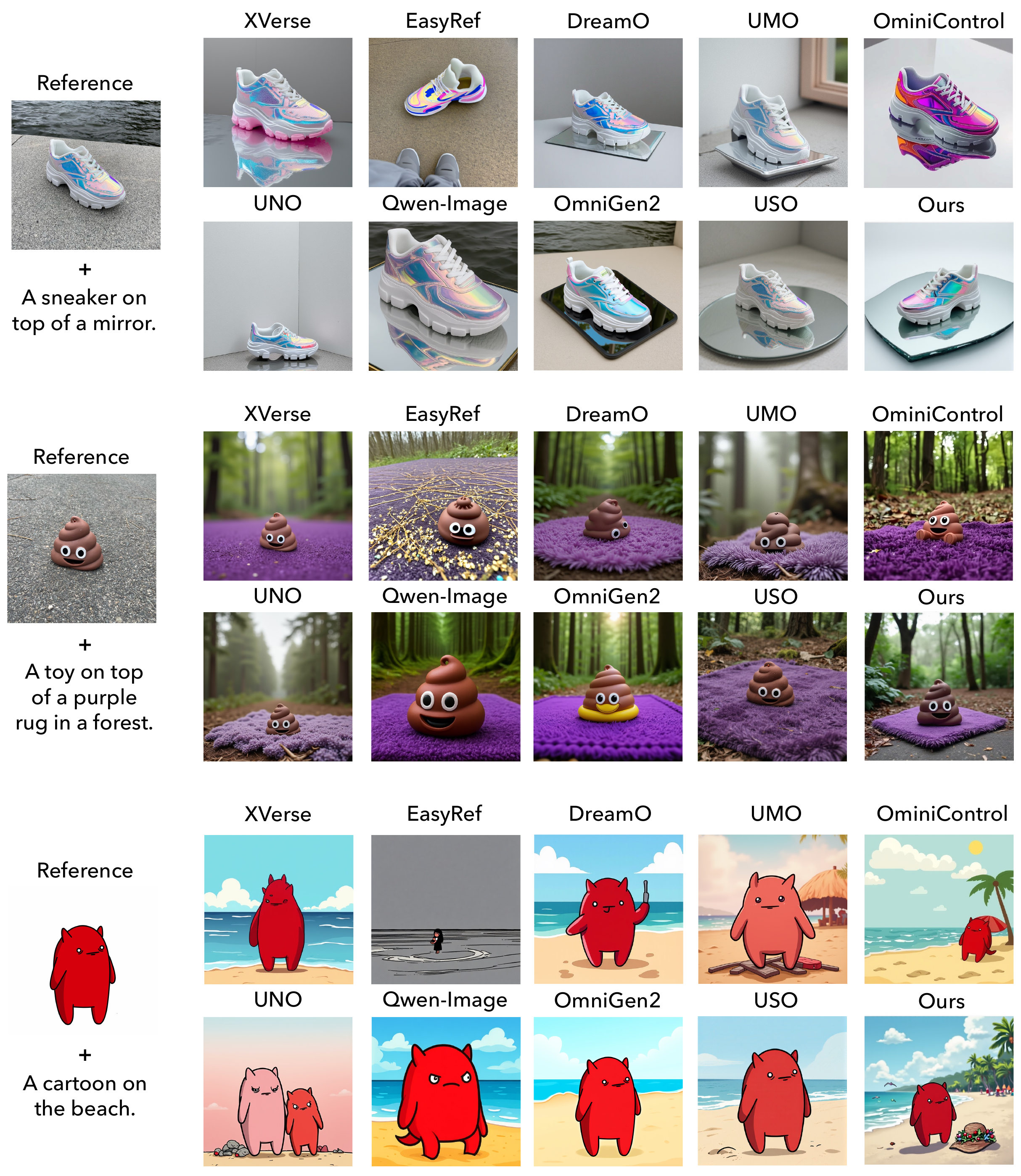}
    \caption{Additional qualitative comparisons with state-of-the-art subject-driven generation methods. Our method consistently achieves better identity preservation and text alignment across diverse prompts. We compare against XVerse~\cite{chen2025xverse}, EasyRef~\cite{zong2024easyref}, DreamO~\cite{mou2025dreamo}, UMO~\cite{cheng2025umo}, OminiControl~\cite{tan2024ominicontrol}, UNO~\cite{wu2025less}, Qwen-Image~\cite{wu2025qwen}, OmniGen2~\cite{wu2025omnigen2}, and USO~\cite{wu2025uso}.}
\label{fig:supp_figure_more_comparison_2}
\end{figure*}

In this section, we provide additional qualitative comparisons with state-of-the-art methods, supplementing the limited space in the main paper. As shown in Figure~\ref{fig:supp_figure_more_comparison_1} and Figure~\ref{fig:supp_figure_more_comparison_2}, we compare our approach with XVerse~\cite{chen2025xverse}, EasyRef~\cite{zong2024easyref}, DreamO~\cite{mou2025dreamo}, UMO~\cite{cheng2025umo}, OminiControl~\cite{tan2024ominicontrol}, UNO~\cite{wu2025less}, Qwen-Image~\cite{wu2025qwen}, OmniGen2~\cite{wu2025omnigen2}, and USO~\cite{wu2025uso}. Our approach achieves competitive subject
identity with diverse subject pose variations, alleviating the copy-paste issue from other VAE-based models.

\section{Licenses for existing assets}
\label{sec:licenses}
The following list contains licenses for data and model used in the paper:

\begin{itemize}
    \item Flux~\cite{blackforestlabs2024flux}: Apache License 2.0
    
    \url{https://github.com/black-forest-labs/flux}
    \item InternVL-3~\cite{zhu2025internvl3}: MIT License
    
    \url{https://github.com/opengvlab/internvl}
    \item UNO-1M~\cite{wu2025less}: Apache License 2.0
    
    \url{https://github.com/bytedance/UNO}
    \item DreamBench~\cite{ruiz2023dreambooth}: CC-BY-4.0 License
    
    \url{https://github.com/google/dreambooth}
    \item XVerseBench~\cite{chen2025xverse}: Apache License 2.0
    
    \url{https://github.com/bytedance/XVerse}
    \item LAMICBench~\cite{chen2025lamic}: Apache License 2.0
    
    \url{https://github.com/Suchenl/LAMIC}
    \item DreamBench++~\cite{peng2025dreambench}: Apache License 2.0

    \url{https://github.com/yuangpeng/dreambench_plus}
\end{itemize}

\section{Discussions and limitations}
\label{sec:discussions}
One of the limitations of the current framework lies in the alignment between the MLLM text representation space and the DiT text conditioning space, which was originally designed to operate with the T5 encoder~\cite{t5_paper}. Pretrained diffusion models such as Flux~\cite{blackforestlabs2024flux} require substantial computational resources and massive text-to-image data to achieve effective alignment between the T5 text encoding space and the DiT conditioning space. In the current case, notably, even with no dedicated text-to-image alignment stage, our approach can already achieve comparable text alignment performance on standard benchmarks, and evidently superior prompt adherence on multimodal understanding. This suggests that, given sufficient computational budget and high-quality text-to-image data, our MLLM-DiT system would likely exhibit improved text-following capabilities.

Another limitation lies in the scope of multi-subject evaluation, although we manifest the model's capability of easily adapting to multi-subject scenarios in Section~\ref{sec:multi_subject}. This is also owing in part to the scarcity nature of high-quality multi-subject data collections. Moreover, as the primary objective of this work is to investigate optimal strategies for squeezing MLLM capacity for subject-driven generation, multi-subject scenarios are less discussed to prevent from getting distracted from our central focus. Nevertheless, studying whether MLLMs can provide benefits with their internal knowledge for multi-subject harmonization and interaction—including, for instance, physical interactions between subjects—can be a promising direction for future research.

\section{Societal impact}
\label{sec:societal_impact}
We expect our work to have a meaningful and positive societal impact by enabling more flexible and accessible personalized image generation. In particular, we sincerely wish that our method can help users express their creativity by generating personalized visual content for versatile applications. Moreover, we hope that our work serves as \textit{the hitchhiker's guide} for future research to be aware of the great benefits of leveraging MLLMs for subject-driven generation, and to explore even more effective solutions to further squeeze capacity from MLLMs for various subject-driven tasks.

\noindent\textbf{Potential negative societal impact.} Our work is likely to be similar as other research on data generation regarding potential negative societal impact with the risk of digital forgery. In addition, unintended or inappropriate use of the technique may raise copyright and ethical concerns.

\end{document}